\documentclass{article}

\usepackage{eccv26}
\usepackage[utf8]{inputenc} % allow utf-8 input
\usepackage[T1]{fontenc}    % use 8-bit T1 fonts
\usepackage{hyperref}       % hyperlinks
\usepackage{url}            % simple URL typesetting
\usepackage{booktabs}       % professional-quality tables
\usepackage{amsfonts}       % blackboard math symbols
\usepackage{nicefrac}       % compact symbols for 1/2, etc.
\usepackage{microtype}      % microtypography
\usepackage{lipsum}
\usepackage{amsmath}
\usepackage{algorithm}
\usepackage{multirow}
\usepackage{fancyhdr}       % header
\usepackage{graphicx}       % graphics
\graphicspath{{media/}}     % organize your images and other figures under media/ folder
\usepackage{amssymb}
\usepackage{algorithmic}
\usepackage{caption}
\usepackage{subcaption}
\usepackage{listings}
\usepackage{microtype}

\newcommand{\etal}{\textit{et al}.}

\title{Spectral-Structured Diffusion for Single-Image Rain Removal
}

\author{
  Yucheng Xing \\
  Department of Electrical and Computer Engineering \\
  Stony Brook University \\
  Stony Brook, NY 11794, USA \\
  \texttt{yucheng.xing@stonybrook.edu} \\
  \And
  Xin Wang \\
  Department of Electrical and Computer Engineering \\
  Stony Brook University \\
  Stony Brook, NY 11794, USA \\
  \texttt{x.wang@stonybrook.edu} \\
}

\begin{document}
\maketitle

%\blfootnote{This is a preprint version of the paper under review.}

%--------------------------------------------------
%       Section 0 : Abstract
%--------------------------------------------------

% ~ 1/4 page

\begin{abstract}

Rain streaks manifest as directional and frequency-concentrated structures that overlap across multiple scales, making single-image rain removal particularly challenging. While diffusion-based restoration models provide a powerful framework for progressive denoising, standard spatial-domain diffusion does not explicitly account for such structured spectral characteristics. We introduce \textbf{\textit{SpectralDiff}}, a spectral-structured diffusion-based framework tailored for single-image rain removal. Rather than redefining the diffusion formulation, our method incorporates structured spectral perturbations to guide the progressive suppression of multi-directional rain components. To support this design, we further propose a full-product U-Net architecture that leverages the convolution theorem to replace convolution operations with element-wise product layers, improving computational efficiency while preserving modeling capacity. Extensive experiments on synthetic and real-world benchmarks demonstrate that SpectralDiff achieves competitive rain removal performance with improved model compactness and favorable inference efficiency compared to existing diffusion-based approaches. 

\end{abstract}

%--------------------------------------------------
%       Section 1 : Introduction
%--------------------------------------------------

% ~ 1 page

%\newpage
\section{Introduction ~\label{sec:1}}

Rain streaks significantly degrade image visibility and interfere with scene understanding in outdoor vision systems. They exhibit strong directional and frequency-concentrated characteristics that overlap across multiple scales, interacting with scene textures and edges. As a result, suppressing rain while preserving fine details remains challenging in the single-image setting. 

Single-image rain removal is fundamentally challenging for several reasons. First, rain streaks possess distinctive physical and spectral structure, yet many existing models do not explicitly encode such properties, often leading to over-smoothing or detail loss. Second, the problem is inherently ill-posed, as the separation between rain patterns and scene details often depends on strong priors that may not generalize to complex real-world conditions. Third, unlike video deraining, single-image settings lack temporal cues, making it difficult to isolate rain layers without introducing artifacts. 

Existing deraining methods span both model-driven and data-driven approaches. Early techniques rely on handcrafted frequency decomposition or layer separation priors~\cite{xu2012improved, ding2016single, zheng2013single, kim2013single}, which often struggle in complex real-world conditions. With the rise of deep learning, CNN-based models directly learn rainy-to-clean mappings~\cite{fu2017clearing, fu2017removing}, while multi-stage~\cite{ren2019progressive, wang2020model, zamir2021multi, chen2021hinet} and recurrent architectures~\cite{li2018recurrent, yang2019single}, progressively remove rain layers. More recently, generative models, including GAN-based~\cite{li2019heavy, wei2021deraincyclegan, zhang2019image, wang2021rain} and AutoEncoder-based~\cite{hu2019depth, li2018non, wang2019erl} approaches, further improve visual quality. However, most of these methods rely on assumptions that may break on real data, and do not explicitly incorporate the structured spectral characteristics of rain streaks into the restoration process. 

Diffusion-based restoration models provide a progressive refinement mechanism that iteratively removes corruption and recover image details. This step-wise denoising process naturally aligns with the layered characteristic of rain streaks, where degradations can be gradually attenuated across multiple stages. Motivated by this observation, we build upon the diffusion  model~\cite{ho2020denoising} and introduce \textit{SpectralDiff}, a spectral-structured diffusion framework tailored for single-image rain removal. Rather than redefining the diffusion formulation, our spectral-structured perturbation design introduces directional and scale-aware modulation guided by rain-specific frequency structure. This design exposes the model to diverse rain patterns across orientations and scales during training, reducing reliance on rigid handcrafted priors and improving robustness under complex real-world conditions. To support this design efficiently, we further propose a full-product U-Net architecture that leverages the convolution theorem to replace convolution operations with equivalent element-wise product layers, improving computational efficiency while preserving modeling capacity. Extensive experiments on synthetic and real-world benchmarks demonstrate that SpectralDiff achieves competitive deraining performance while offering improved model compactness and favorable inference efficiency compared to existing diffusion-based approaches.

Our contributions are summarized as follows:
\begin{itemize}
    \item We introduce SpectralDiff, a spectral-structured diffusion framework tailored for single-image rain removal. 
    \item We design structured spectral perturbations that incorporate rain-specific frequency characteristics into the diffusion-based restoration process.
    \item We propose a full-product U-Net architecture that improves computational efficiency through operator-level redesign while preserving modeling capacity. 
\end{itemize} 

The remainder of this paper is organized as follows: Sec.~\ref{sec:2} reviews related works, while the details of our model are provided in Sec.~\ref{sec:3}. To evaluate its effectiveness, we have conducted extensive experiments as well as ablation studies in Sec.~\ref{sec:4}. Finally, we make a conclusion and point out potential future directions in Sec.~\ref{sec:5}. 
Code will be released upon acceptance.  

%--------------------------------------------------
%       Section 2 : Related Works
%--------------------------------------------------

% ~ 1 page

%\newpage
\section{Related Works~\label{sec:2}}

%%%% Section 2.1 : Single Image Rain Removal
%%%%%%%%%%%%%%%%%%%%%%%%%%%%%%%%%%%%%%%%%%%%
\subsection{Single Image Rain Removal~\label{sec:2.1}}

Single-image deraining plays an important role in adverse-weather vision systems. Unlike video deraining, single-image rain removal lacks temporal information, making the problem significantly more challenging. Traditional approaches primarily relied on decomposition techniques~\cite{chen2017error, kang2011automatic, luo2015removing} to separate high-frequency rain streaks from background contents~\cite{xu2012removing, zheng2013single, ding2016single, kim2013single}. However, these methods often led to over-smoothing, inadvertently removing fine details in the underlying image. Another line of work introduced prior assumptions about rain, such as rankness, sparsity, directions, and scales, into decomposition models~\cite{li2016rain, chen2014visual, kang2011automatic, zhu2017joint} to improve robustness. 

With the development of deep learning techniques, Fu~\etal~\cite{fu2017clearing} first introduced Convolutional Neural Networks (CNNs) for single image rain removal, inspiring more advanced CNN architectures~\cite{fu2017removing, yang2019joint}. Multi-scale fusion strategies~\cite{jiang2020multi, wang2020dcsfn, jiang2021multi, zhou2024dual} were proposed to capture texture details at different resolutions. In parallel, multi-stage progressive designs ~\cite{ren2019progressive, zamir2021multi, wang2020model, chen2021hinet, das2023rain} stacked similar modules to remove rain streaks layer by layer, while Recurrent Neural Networks (RNNs)~\cite{li2018recurrent, yang2019single} followed a similar progressive philosophy.

More recently, transformer-based models~\cite{zamir2022restormer, guo2023sky, chen2023learning, zhou2024dual, wang2022uformer, xiao2022image} have been adopted for single-image rain removal, leveraging self-attention mechanisms to capture long-range dependencies and improve restoration quality. However, most of these methods do not explicitly encode the spectral structure of rain streaks. Generative approaches, including Generative Adversarial Networks (GANs)~\cite{li2019heavy, wei2021deraincyclegan, zhang2019image, wang2021rain} and AutoEncoders~\cite{hu2019depth, li2018non, wang2019erl}, further enhance perceptual quality by modeling the rain formation process.

Diffusion models have recently been explored for image restoration tasks due to their progressive denoising formulation. Their iterative refinement mechanism provides a natural way to gradually suppress corruption and recover image details. However, their application to single-image deraining remains relatively limited, and existing approaches typically adopt standard spatial Gaussian perturbations~\cite{ozdenizci2023restoring, wei2023raindiffusion, zeng2024multi}. In this work, we investigate how diffusion models can be adapted for deraining by explicitly incorporating rain-specific spectral structure into the denoising trajectory.

%%%% Section 2.2 : Spectral Domain Convolution
%%%%%%%%%%%%%%%%%%%%%%%%%%%%%%%%%%%%%%%%
\subsection{Spectral Domain Convolution~\label{sec:2.3}}

Spectral-domain convolution has been explored to improve both computational efficiency and modeling capacity in deep neural networks. In the spatial domain, convolution requires repeated element-wise multiplications and summations, leading to substantial computational cost, particularly for large inputs or deep architectures. By the convolution theorem~\cite{oppenheim1999discrete}, spatial convolution can be reformulated as element-wise multiplication in the frequency domain via Fast Fourier Transform (FFT), reducing computational complexity and enabling more efficient implementations. Mathieu~\etal~\cite{mathieu2013fast} systematically explored frequency-domain operations in deep networks and reported significant speedups for large kernels and inputs. Similar acceleration effects were further analyzed by Lavin and Gray~\cite{lavin2016fast}. Rippel~\etal~\cite{rippel2015spectral} introduced spectral pooling and spectral parameterization, demonstrating that spectral representations can improve both efficiency and training behavior.

Subsequent work extended spectral techniques beyond acceleration. Cheng~\etal~\cite{cheng2015exploration} approximated convolution kernels as circulant matrices to exploit their diagonalization in the frequency domain. Chi~\etal~\cite{chi2019fast} proposed spectral residual learning to model global dependencies efficiently. Spectral operations have also been integrated into modern architectures and optimized within widely used libraries~\cite{vasilache2014fast, nvidia_cufft}.

Beyond computational benefits, frequency-domain representations have shown advantages in modeling structured patterns. Zhong~\etal~\cite{zhong2018joint} employed wavelet-based representations for high-quality image restoration, while Chi~\etal~\cite{chi2020fast} introduced hybrid spatial-spectral units to achieve mixed receptive fields. Such hybrid designs have been adopted in various applications~\cite{shchekotov2022ffc, chu2023rethinking, chen2024globalsr}. As noted in \cite{shchekotov2022ffc}, spectral operations are particularly effective in capturing periodic or quasi-periodic structures, which are often observed in rain streak patterns. This observation motivates the incorporation of spectral modeling into our deraining framework.

%--------------------------------------------------
%       Section 3 : Methodology
%--------------------------------------------------

% ~ 2 page

\section{Methodology~\label{sec:3}}

In this section, we analyze the relationship between the layered structure of rain streaks and the iterative denoising mechanism of diffusion models. Building upon this observation, we present \textit{SpectralDiff}, a spectral-structured diffusion framework that progressively suppresses rain components via a structured spectral perturbation design. We further introduce an operator-level redesign of the U-Net architecture to improve computational efficiency. Together, these components define a spectrally structured and computationally efficient framework for single-image rain removal.

%%%% Section 3.1 : Layer Characteristic
%%%%%%%%%%%%%%%%%%%%%%%%%%%%%%%%%%%%%%%
\subsection{Layered Characteristic of Rain Streaks~\label{sec:3.1}}

\begin{figure*}[!htpb]
    \centering
    \begin{subfigure}{.19\linewidth}
        \centering
        \includegraphics[width=\linewidth]{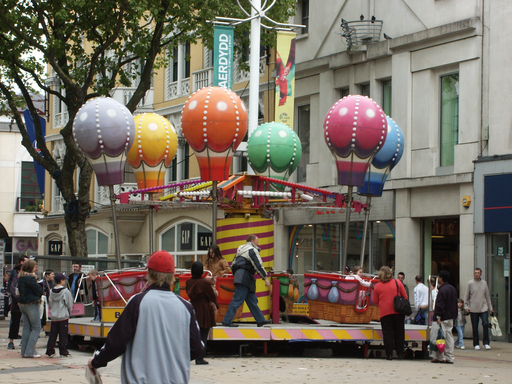}
        \subcaption[]{}
        \label{fig:rain_process_a1}
    \end{subfigure}
    \hfill
    \begin{subfigure}{.19\linewidth}
        \centering
        \includegraphics[width=\linewidth]{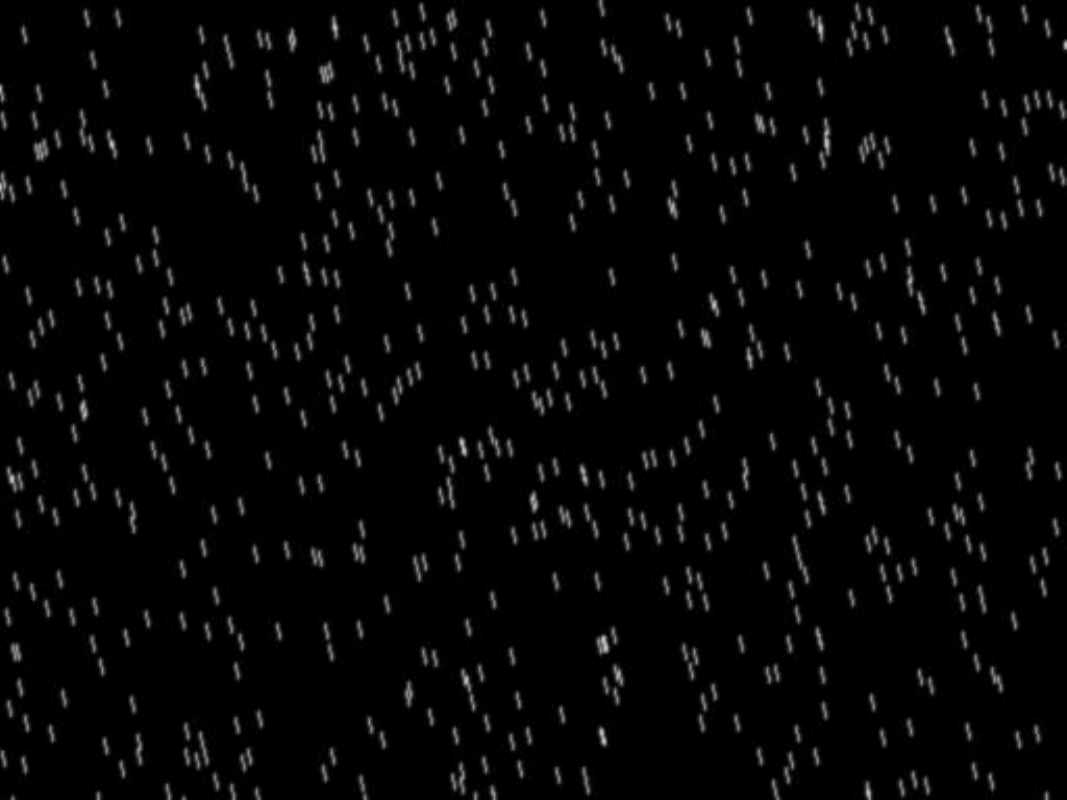}
        \subcaption[]{}
        \label{fig:rain_process_b1}
    \end{subfigure}
    \hfill
    \begin{subfigure}{.19\linewidth}
        \centering
        \includegraphics[width=\linewidth]{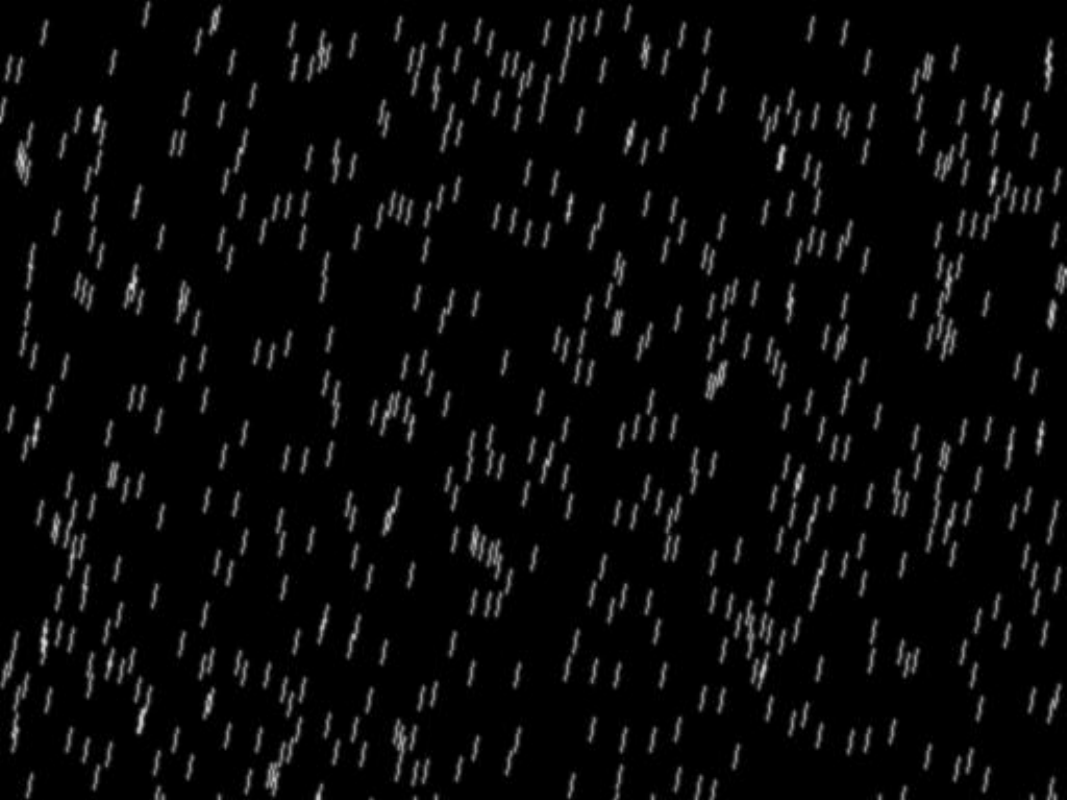}
        \subcaption[]{}
        \label{fig:rain_process_c1}
    \end{subfigure}
    \hfill
    \begin{subfigure}{.19\linewidth}
        \centering
        \includegraphics[width=\linewidth]{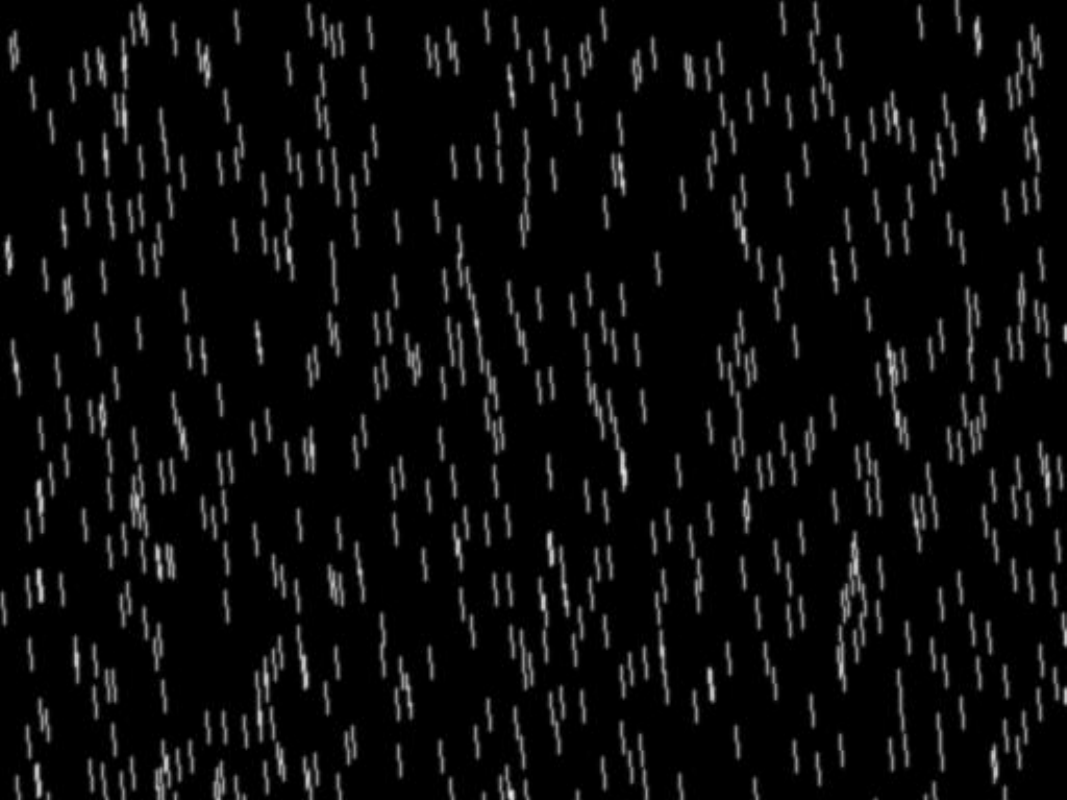}
        \subcaption[]{}
        \label{fig:rain_process_d1}
    \end{subfigure}
    \hfill
    %\begin{subfigure}{.15\linewidth}
    %    \centering
    %    \includegraphics[width=\linewidth]{Fig/blurred_rain_mask_4.pdf}
    %    \subcaption[]{}
    %    \label{fig:rain_process_e1}
    %\end{subfigure}
    %\hfill
    \begin{subfigure}{.19\linewidth}
        \centering
        \includegraphics[width=\linewidth]{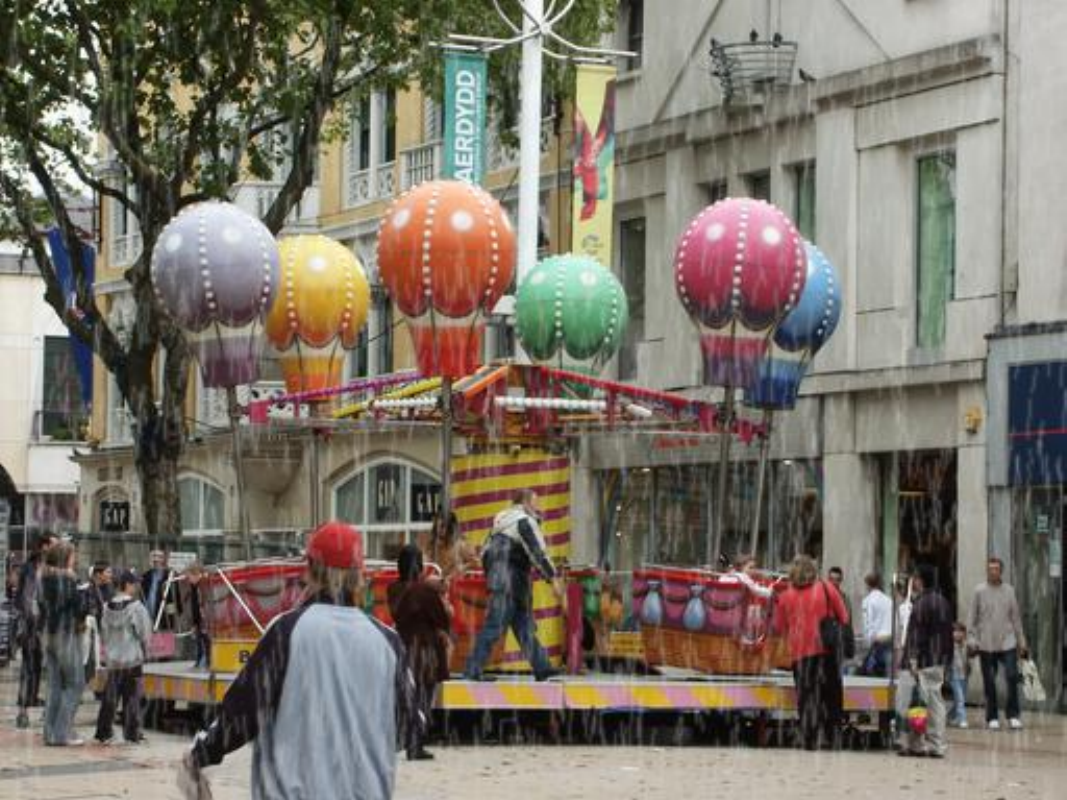}
        \subcaption[]{}
        \label{fig:rain_process_f1}
    \end{subfigure}
    %\vspace{-0.1in}
    \caption{Illustration of the rain layer adding process. \ref{fig:rain_process_a1} is the original image $\textbf{B}$, \ref{fig:rain_process_b1} - \ref{fig:rain_process_d1} are example rain-streak layers $\textbf{R}_{d}$ $(d \in [1, 3])$, and \ref{fig:rain_process_f1} is the final rainy image $\textbf{O}$.}
    \label{fig:rain_process}
    %\vspace{-0.2in}
\end{figure*}

Given a rainy image $\mathbf{O}$, it can be modeled as the superposition of a clear image $\mathbf{B}$ with background content and a mask $\textbf{R}$ of rain streaks, which can be expressed as
\begin{equation}
    %\small
    \mathbf{O} = \mathbf{B} + \mathbf{R}.~\label{eq:rain_streak}
\end{equation}
In real-world scenarios, rain streaks exhibit variations in scale and direction due to factors such as rain density and wind speed. Based on this observation, we can divide the rain-streak mask $\mathbf{R}$ into multiple layers, where each layer consists of streaks sharing similar physical properties (e.g., scale, thickness and direction), as illustrated in Fig.~\ref{fig:rain_process}. Accordingly, we can rewrite Eq.~\ref{eq:rain_streak} as
\begin{equation}
    %\small
    \mathbf{O} = \mathbf{B} + \sum^{D}_{d}\mathbf{R}_{d},~\label{eq:rain_streaks}
\end{equation}
where $\textbf{R}_{d}$ represents the mask corresponding to the $d$-th rain layer, which contains linear rain streak components, and $D$ denotes the total number of such rain streak layers. While some prior multi-stage methods have attempted to model this layered structure, they remain limited in effectiveness. In practice, the true value of $D$ in real-world rainy scenes can be significantly larger and more complex than the number of stages pre-defined in these models, making such approximations insufficient for fully capturing the diverse rain streak patterns.

\begin{figure}[!htpb]
    \centering
    \includegraphics[width=0.8\linewidth]{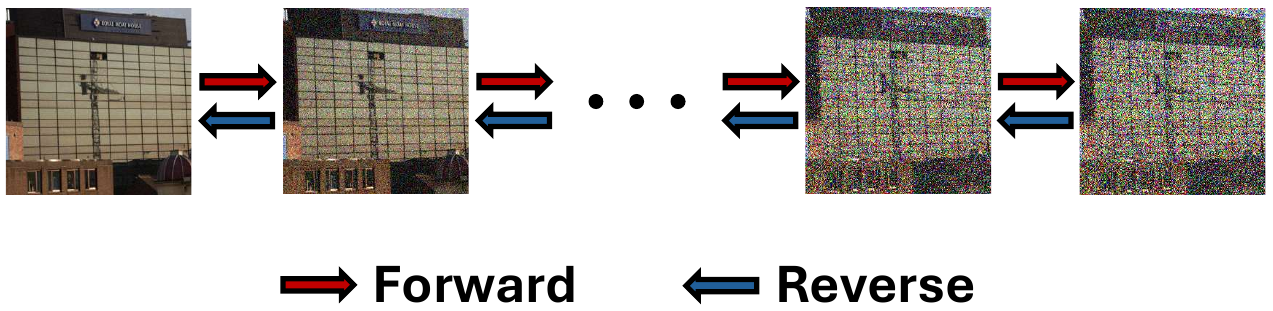}
    %\vspace{-0.1in}
    \caption{Illustration of forward / reverse diffusion process.}
    \label{fig:diffusion_process}
    %\vspace{-0.2in}
\end{figure}

Diffusion models, due to their iterative denoising mechanism, naturally provide a progressive transformation framework that conceptually aligns with such layered structures. In standard diffusion models, a clean image $\mathbf{x}_{0}$ is progressively transformed into a pure Gaussian noise mask $\mathbf{x}_{T}$ by iteratively adding Gaussian noise $\mathbf{\epsilon}\sim\mathcal{N}(\mathbf{0}, \mathbf{I})$ through a Markov chain, and this diffusion process can be formulated as
\begin{equation}
    %\small
    \mathbf{x}_{t} = \sqrt{1 - \beta_{t}}\mathbf{x}_{t-1} + \sqrt{\beta_{t}}
     \mathbf{\epsilon},~\label{eq:diff_forward1}
\end{equation}
i.e.,
\begin{equation}
    %\small
    q(\mathbf{x}_{t}|\mathbf{x}_{t-1}) = \mathcal{N}(\mathbf{x}_{t}; \sqrt{1 - \beta_{t}}\mathbf{x}_{t-1}, \beta_{t}\mathbf{I}), ~\label{eq:diff_forward2}
\end{equation}
where $T$ is the total number of steps in this process and $\{\beta_{1}, \dots, \beta_{T}\}$ is the corresponding noise variance schedule. To obtain a clear and realistic image, starting from a pure noise $p(\mathbf{x}_{T}) = \mathcal{N}(\mathbf{x}_{T}; \mathbf{0}, \mathbf{I})$, the diffusion process in Eqs.~\eqref{eq:diff_forward1}-\eqref{eq:diff_forward2} is reversed with the learned transitions
\begin{equation}
    %\small
    p_{\phi}(\mathbf{x}_{t-1}|\mathbf{x}_{t}) = \mathcal{N}(\mathbf{x}_{t-1}; \mu_{\phi}(\mathbf{x}_{t}, t), \beta_{t}\mathbf{I}),~\label{eq:diff_backward}
\end{equation}
where $\mu_{\phi}(\cdot)$ is parameterized by a neural network. Both forward and reverse diffusion processes are illustrated in Fig.~\ref{fig:diffusion_process}.

By comparing Fig.~\ref{fig:rain_process} and Fig.~\ref{fig:diffusion_process}, as well as Eq.~\ref{eq:rain_streaks} and Eq.~\ref{eq:diff_forward1}, we observe a conceptual resemblance between the layered accumulation of rain streaks and the iterative structure of diffusion models. In both cases, complex observations are formed or refined through a sequence of incremental transformations. This structural analogy suggests that the removal of rain may benefit from a progressive restoration strategy, where different rain components are attenuated across multiple steps rather than in a single pass. In particular, if rainy images can be viewed as the superposition of multiple rain layers with diverse orientations and scales, then a step-wise denoising process naturally aligns with this layered structure. For notational convenience, we associate the diffusion step index with conceptual rain layers, allowing us to describe the restoration process in a unified progressive framework. This alignment serves as an intuitive bridge between the physical decomposition in Eq.~\ref{eq:rain_streaks} and the iterative formulation of diffusion models.

%%%% Section 3.2 : Spectral Diffusion
%%%%%%%%%%%%%%%%%%%%%%%%%%%%%%%%%%%%%
\subsection{Iterative Rain Removal via Spectral Diffusion~\label{sec:3.2}}

Inspired by the layered interpretation in Sec.~\ref{sec:3.1}, we adopt diffusion as a progressive restoration mechanism for single-image rain removal. While diffusion models have shown strong performance in conditional generation and restoration tasks, most existing approaches~\cite{ozdenizci2023restoring, wei2023raindiffusion, zeng2024multi} employ standard isotropic Gaussian perturbations in the spatial domain. However, rain residuals are neither spatially independent nor isotropic. Each rain layer $\textbf{R}_{d}$ consists of elongated, direction- and scale-consistent structures with correlated spatial patterns, which are not well represented by spatial i.i.d. Gaussian noise. 

The key observation is that although rain streaks are highly structured in the spatial domain, their dominant frequency content can be compactly characterized in the spectral domain. Straight and elongated structures correspond to concentrated energy ridges in the spectral domain, oriented orthogonally to their spatial direction. Fine and dense streaks mainly occupy higher frequency bands, while thicker streaks are associated with lower or medium frequencies. Therefore, modeling the dominant frequency distribution of rain layers is often more critical than enforcing a specific spatial marginal distribution. 

\begin{figure}[!htpb]
    \centering
    \begin{subfigure}{\linewidth}
        \centering
        \includegraphics[width=0.24\linewidth]{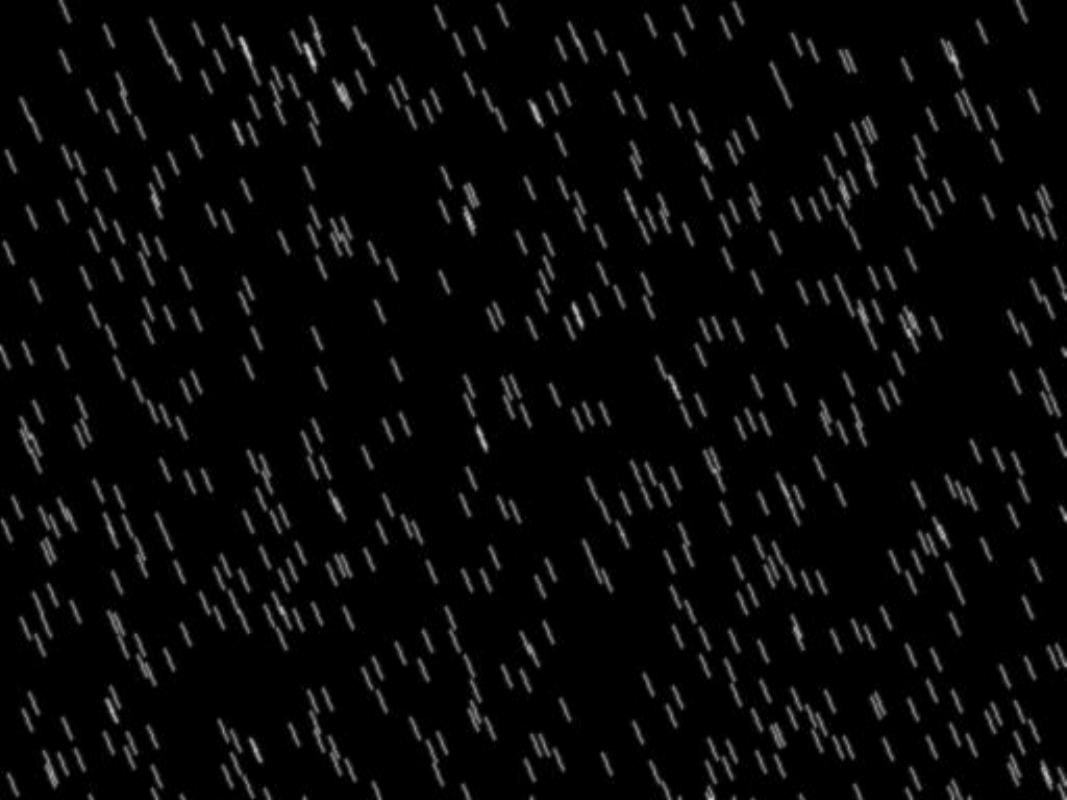}
        \includegraphics[width=0.24\linewidth]{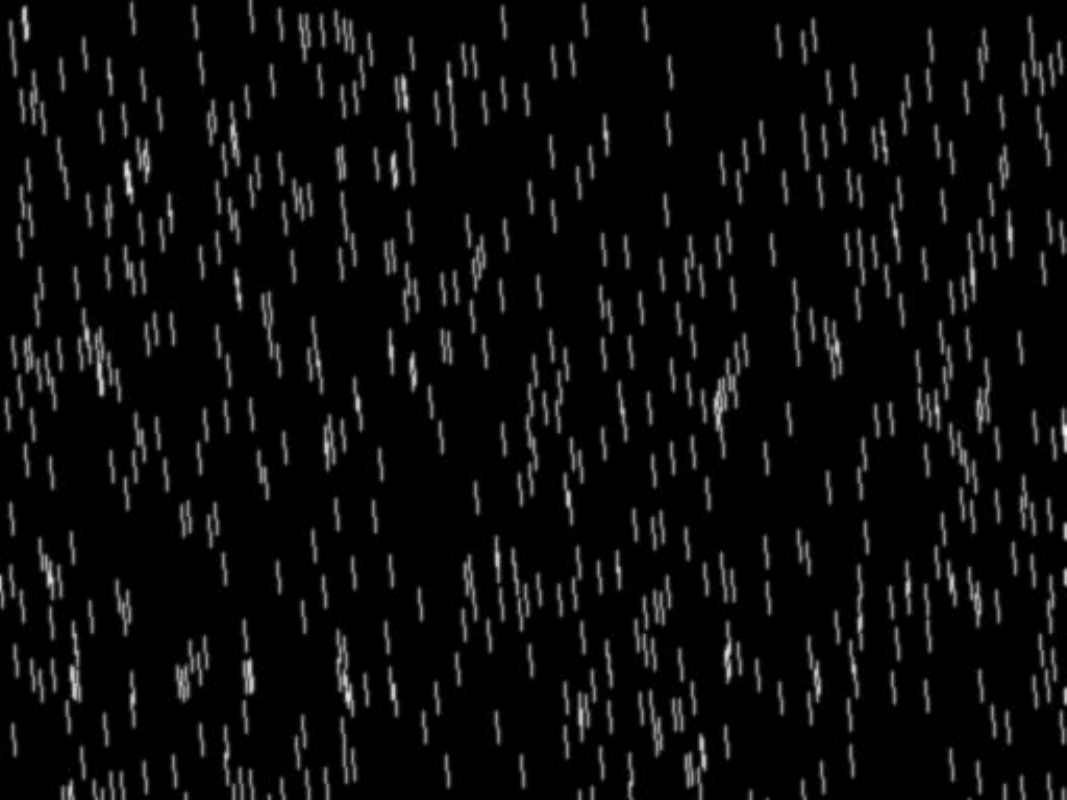}
        \includegraphics[width=0.24\linewidth]{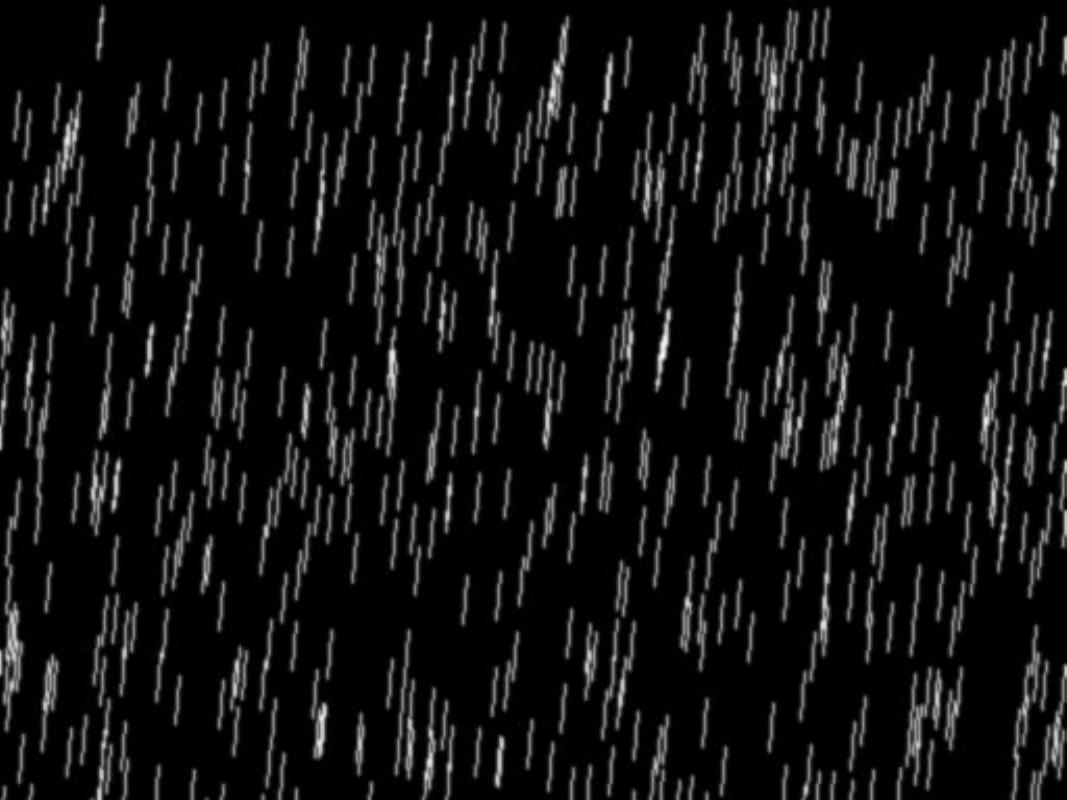}
        \includegraphics[width=0.24\linewidth]{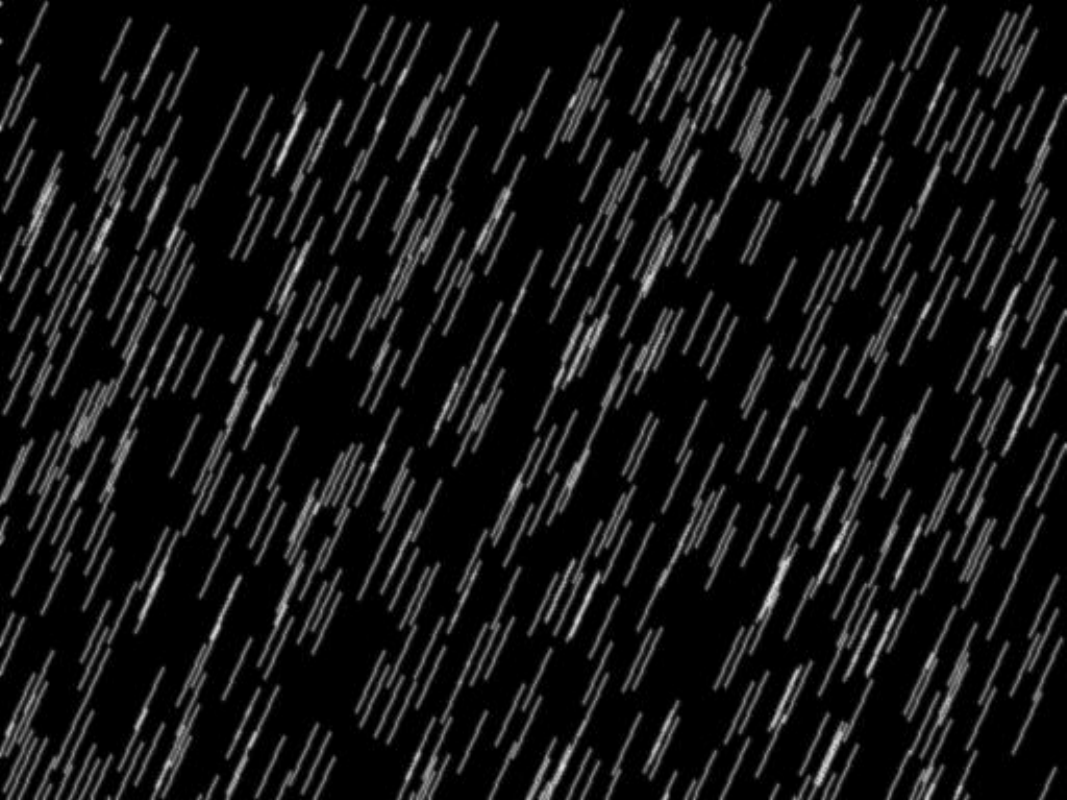}
        %%\vspace{-0.03in}
        \subcaption[]{Rain-Streak Layers $\mathbf{R}_{d}$}
        \label{fig:rain_process_a}
    \end{subfigure}
    \hfill
    \begin{subfigure}{\linewidth}
        \centering
        \includegraphics[width=0.24\linewidth]{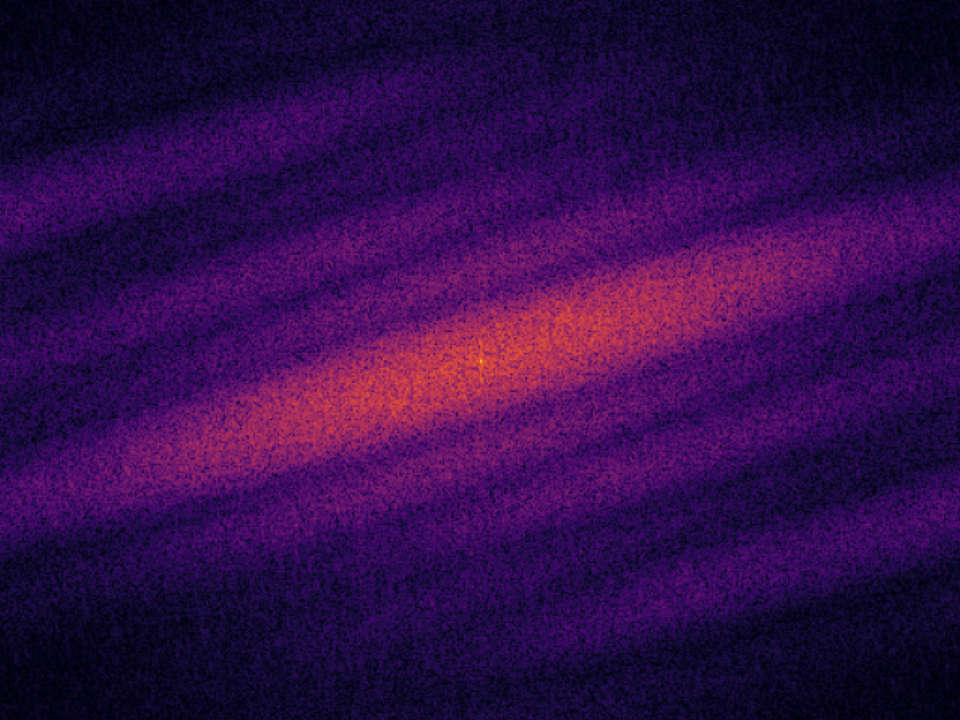}
        \includegraphics[width=0.24\linewidth]{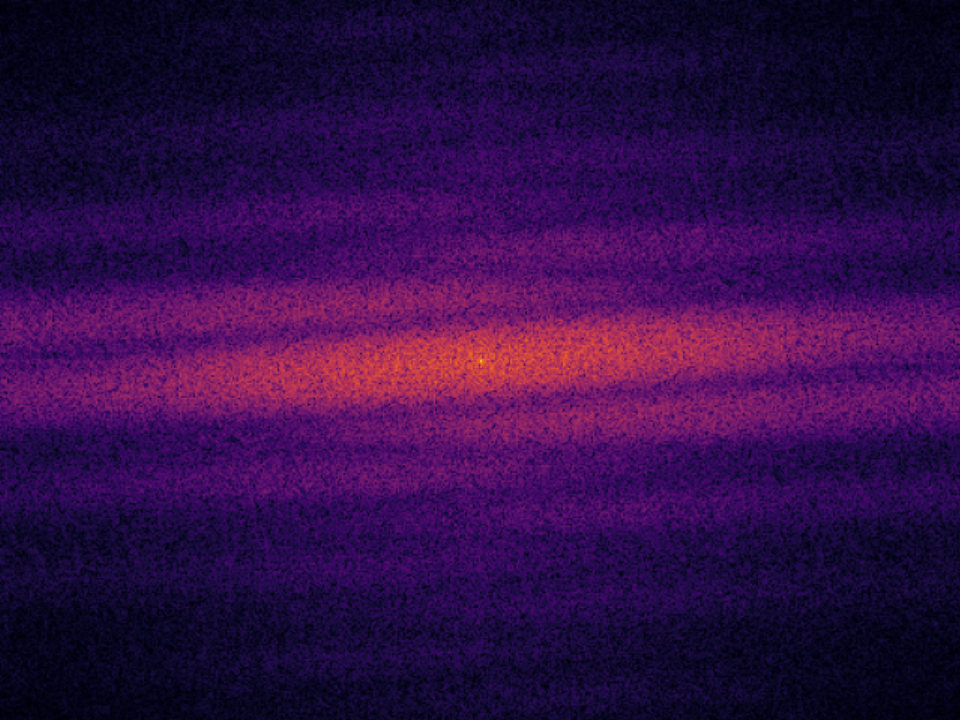}
        \includegraphics[width=0.24\linewidth]{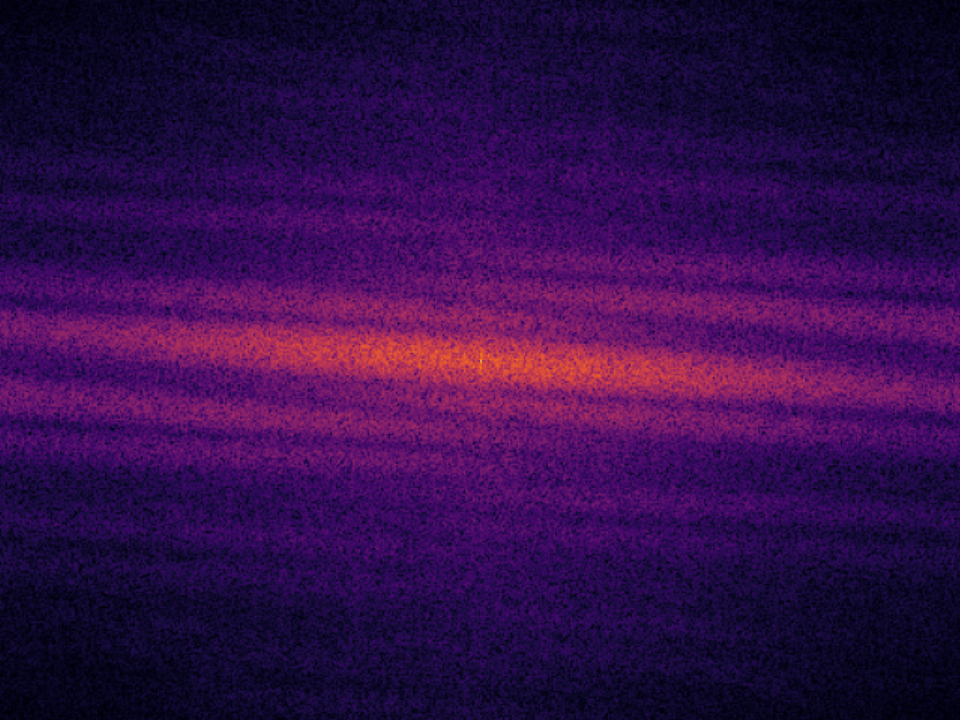}
        \includegraphics[width=0.24\linewidth]{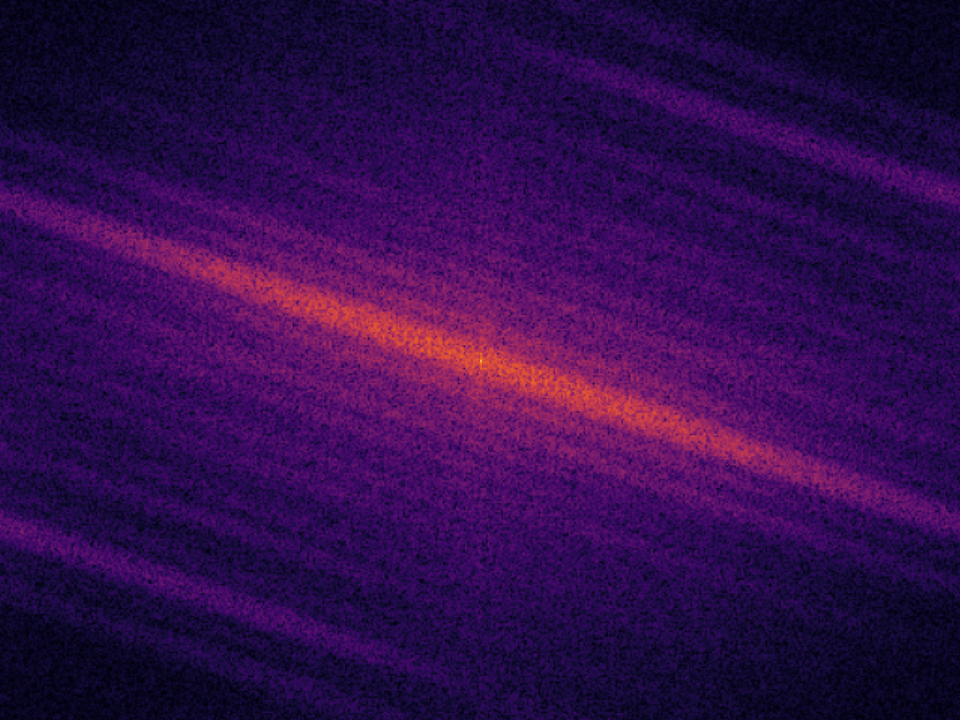}
        %%\vspace{-0.03in}
        \subcaption[]{Magnitude Map of $\mathcal{F}(\mathbf{R}_{d})$}
        \label{fig:rain_process_b}
    \end{subfigure}
    \hfill
    \begin{subfigure}{\linewidth}
        \centering
        \includegraphics[width=0.24\linewidth]{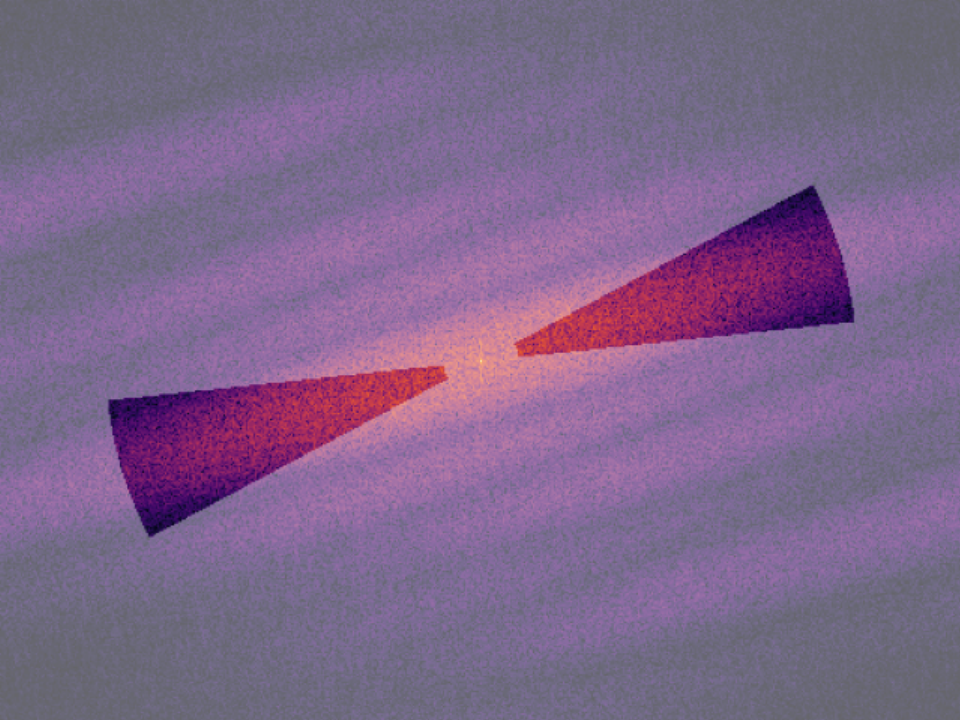}
        \includegraphics[width=0.24\linewidth]{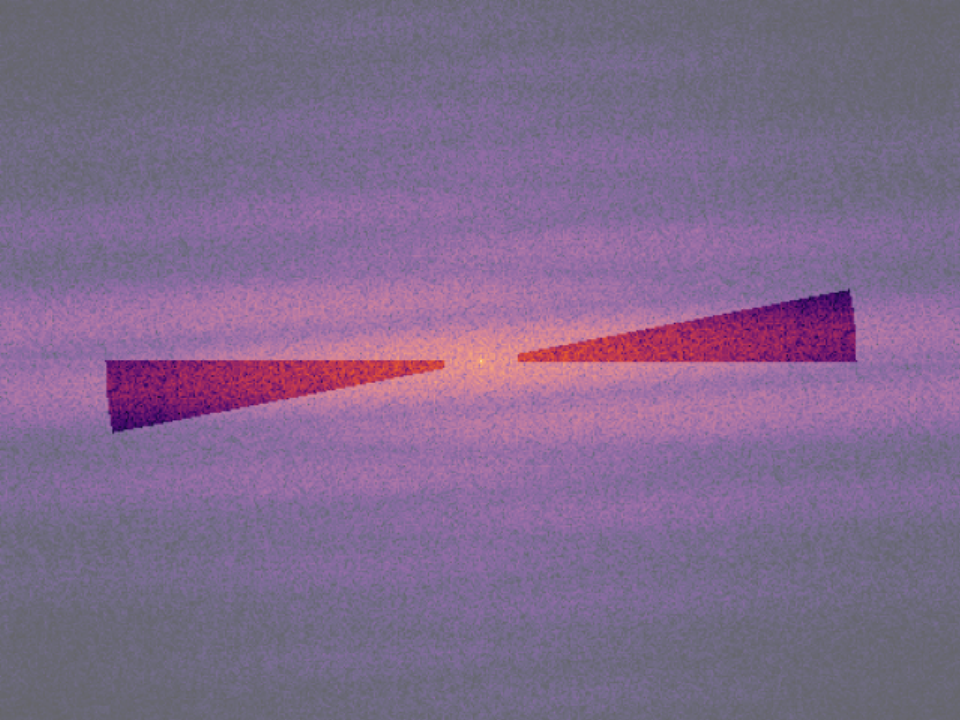}
        \includegraphics[width=0.24\linewidth]{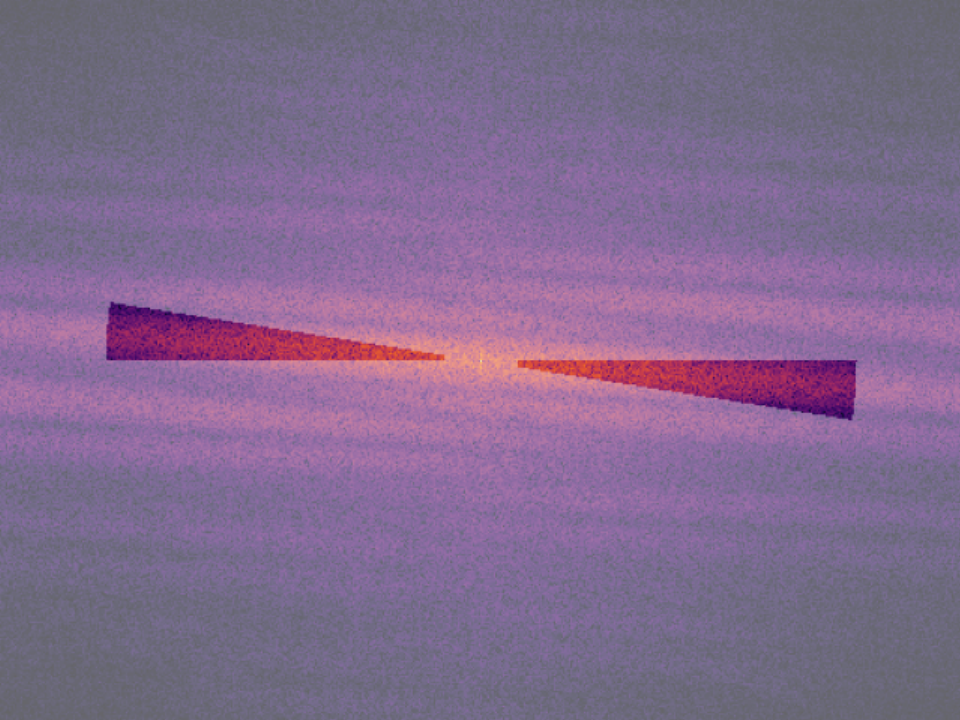}
        \includegraphics[width=0.24\linewidth]{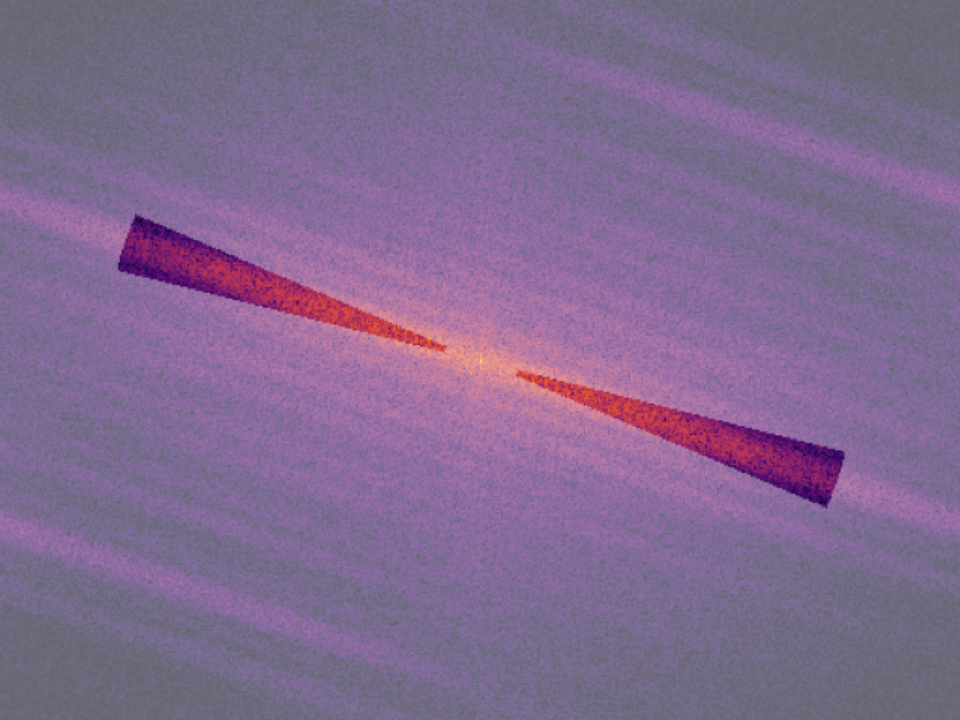}
        %%\vspace{-0.03in}
        \subcaption[]{Magnitude Map of $\mathcal{F}(\mathbf{R}_{d})$ + Directional Sector}
        \label{fig:rain_process_c}
    \end{subfigure}
    %\vspace{-0.1in}
    \caption{Illustration of rain layers in the frequency domain.}
    \label{fig:rain_frequency}
    %\vspace{-0.2in}
\end{figure}

Based on this insight, we introduce a spectral-structured perturbation design. Instead of injecting isotropic spatial noise, we modulate complex Gaussian noise in the frequency domain using direction- and scale-aware masks, as illustrated in Fig.~\ref{fig:rain_frequency}. This produces structured spectral perturbations whose frequency matches the dominant frequency characteristics of rain streak layers. Formally, we use the following structured perturbation proxy:
\begin{equation}
    %\small
    \mathbf{R}_{d} \simeq\mathcal{F}^{-1}(\mathbf{M}_{d}\odot\mathbf{\epsilon}_{f}),~\label{eq:modulation}
\end{equation}
where $\mathbf{\epsilon}_{f} \sim \mathcal{N}_{\mathbb{C}}(\mathbf{0}, \mathbf{I})$ denotes i.i.d. complex Gaussian noise (i.e., $\mathbf{\epsilon}_{f} = \mathbf{\epsilon}_{f, \text{real}} + j\mathbf{\epsilon}_{f, \text{imag}}$ with $\mathbf{\epsilon}_{f, \text{real}}, \mathbf{\epsilon}_{f, \text{imag}} \sim \mathcal{N}(\mathbf{0}, \mathbf{I})$), $\mathbf{M}_{d}$ is a frequency-direction-aware mask associated with rain layer $\mathbf{R}_{d}$, and $\mathcal{F}^{-1}(\cdot)$ denotes the inverse FFT (IFFT). We emphasize that this formulation serves as a structured spectral perturbation mechanism capturing dominant frequency, rather than an exact physical rain simulator.   

For each spectral coordinate $\mathbf{f} = \{f_{x}, f_{y}\}$, the mask $\mathbf{M}_{d}$ determines how noise energy is distributed. It consists of radial and angular components:

\paragraph{\textbf{Radial Mask $\mathbf{M}_{d, r}$.}}
Rain streaks within a layer share similar scale and density, corresponding to specific radial frequency bands. We model this using a Gaussian-shaped band-pass filter: 
\begin{equation}
    %\small
    \mathbf{M}_{d, r} = \mathbf{M}_{d, r}^{r_{d}, \sigma_{r_{d}}}(\mathbf{f}) = \exp{\left(-\frac{(r - r_{d})^{2}}{2\sigma_{r_{d}}^{2}}\right)},~\label{eq:Mr}
\end{equation}
where $r = \left\lVert\mathbf{f}\right\rVert_{2} = \sqrt{f_{x}^{2} + f_{y}^{2}}$ is the radial frequency, $r_{d}$ denotes the density of rain streaks to be retained in the $d$-th layer by controlling the dominant band, and $\sigma_{r_{d}}$ determines the bandwidth to control the thickness. 

\paragraph{\textbf{Angular Mask $\mathbf{M}_{d, \theta}$.}} Since rain streaks exhibit consistent orientation within each layer, their spectral responses concentrate along specific angular sectors. We model this using a von Mises distribution~\cite{mardia2009directional}:
\begin{equation}
    %\small
    \mathbf{M}_{d, \theta} = \mathbf{M}_{d, \theta}^{\theta_{d}, \kappa_{d}}(\mathbf{f}) = \exp{\left(\kappa_{d}\cos{\left(\theta - \theta_{d}\right)}\right)},~\label{eq:Mtheta}
\end{equation}
where $\theta = \arctan2(f_{y}, f_{x})$ denotes the spectral angle, $\theta_{d}$ is the dominant orientation that is orthogonal to the rain direction, and $\kappa_{d}$ controls angular concentration.

The complete mask is given by 
\begin{equation}
    %\small
    \mathbf{M}_{d}^{r_{d}, \sigma_{r_d}, \theta_{d}, \kappa_{d}}(\mathbf{f}) = \mathbf{M}_{d, r}^{r_{d}, \sigma_{r_d}}(\mathbf{f})\odot\mathbf{M}_{d, \theta}^{\theta_{d}, \kappa_{d}}(\mathbf{f}),~\label{eq:M_Mr_Mtheta}
\end{equation}
and is normalized as
\begin{equation}
    %\small
    \mathbf{M}_{d} = \frac{\sqrt{\mathbf{M}_{d}^{r_{d}, \sigma_{r_d}, \theta_{d}, \kappa_{d}}(\mathbf{f})}}{\left\lVert\sqrt{\mathbf{M}_{d}^{r_{d}, \sigma_{r_d}, \theta_{d}, \kappa_{d}}(\mathbf{f})}\right\rVert_{2}}.~\label{eq:M_norm}
\end{equation}

We now define the spectral-structured diffusion process. Let $\mathbf{x}_{f, 0} = \mathcal{F}(\mathbf{B})$ be the spectral representation of the clean image. The spectral representation of the rainy image $\mathbf{x}_{f, D} = \mathcal{F}(\mathbf{O})$ is obtained by progressively injecting masked complex Gaussian noise:
\begin{equation}
    %\small
    \mathbf{x}_{f, d} = \sqrt{1 - \beta_{d}}\mathbf{x}_{f, d - 1} + \sqrt{\beta_{d}}(\mathbf{M}_{d}\odot\mathbf{\epsilon}_{f}),~\label{eq:spectral_diff_forward_1}
\end{equation}
and we define the corresponding forward transition kernel as
\begin{equation}
    %\small
    q(\mathbf{x}_{f, d} | \mathbf{x}_{f, d - 1}) \triangleq \mathcal{N}_{\mathbb{C}}(\mathbf{x}_{f, d}; \sqrt{1 - \beta_{d}}\mathbf{x}_{f, d - 1}, \beta_{d}\mathbf{M}_{d}^{2}\mathbf{I}).~\label{eq:spectral_diff_forward_2}
\end{equation}
In addition to the standard noise schedule $\beta_{d}$, we introduce layer-wise parameters $(r_{d}, \sigma_{r_d}, \theta_{d}, \kappa_{d})$ that are uniformly sampled to cover diverse rain scales and directions. These parameters are globally defined and fixed across datasets. In practice, this corresponds to pre-computing a fixed mask $\mathbf{M}_d$ for each step $d$, analogous to using a fixed $\beta_d$ schedule.

Because the mask $\mathbf{M}_{d}$ is normalized such that $\sum_{\mathbf{f}}{|\mathbf{M}_{d}|^{2}}=1$, the per-step perturbation maintains controlled second-order energy. Although each step introduces anisotropic spectral perturbations, this normalization ensures stable variance scaling across diffusion steps. Importantly, we do not claim that this masked perturbation is theoretically equivalent to the standard isotropic Gaussian diffusion process. Instead, we treat it as a task-aligned structured corruption mechanism that injects rain-specific spectral structure while retaining a tractable Gaussian parameterization. Empirically, we observe stable optimization and consistent deraining behavior.  

The reverse process is parameterized as
\begin{equation}
    %\small
    p_{\phi}(\mathbf{x}_{f, d - 1}|\mathbf{x}_{f, d}) \triangleq \mathcal{N}_{\mathbb{C}}(\mathbf{x}_{f, d - 1}; \mu_{\phi}(\mathbf{x}_{f, d}, d, \mathbf{c}), \beta_{d}\mathbf{M}_{d}^{2}\mathbf{I}),~\label{eq:spectral_diff_backward}
\end{equation}
where $\mu_{\phi}(\cdot)$ is implemented by a neural network trained to predict the clean spectral representation from $\mathbf{x}_{f, d}$ at step $d$ under condition $\mathbf{c}$. Here, the Gaussian form captures the mean and energy structure induced by the masked perturbation and provides a practical denoising parameterization. It should be interpreted as a modeling choice under the structured corruption, rather than a strict posterior derived from the forward process. Through this spectral-structured diffusion process, rain components are progressively attenuated in a frequency-aware manner, enabling stable and structured single-image rain removal.

%%%% Section 3.3 : Product U-Net
%%%%%%%%%%%%%%%%%%%%%%%%%%%%%%%%%%%%%
\subsection{Accelerating with Full-Product U-Net~\label{sec:3.3}}

While Sec.~\ref{sec:3.2} defines a spectral-structured diffusion process, directly performing denoising in the spectral domain is computationally expensive. A naive implementation requires applying FFT before each reverse step and IFFT afterward, introducing additional transformations. Moreover, spectral representations are complex-valued, doubling the effective computational cost at each diffusion step. Given the inherently iterative nature of diffusion models, such overhead becomes substantial. 

\begin{figure*}[!htpb]
    %%\vspace{-0.2in}
    \centering
    \includegraphics[width=\linewidth]{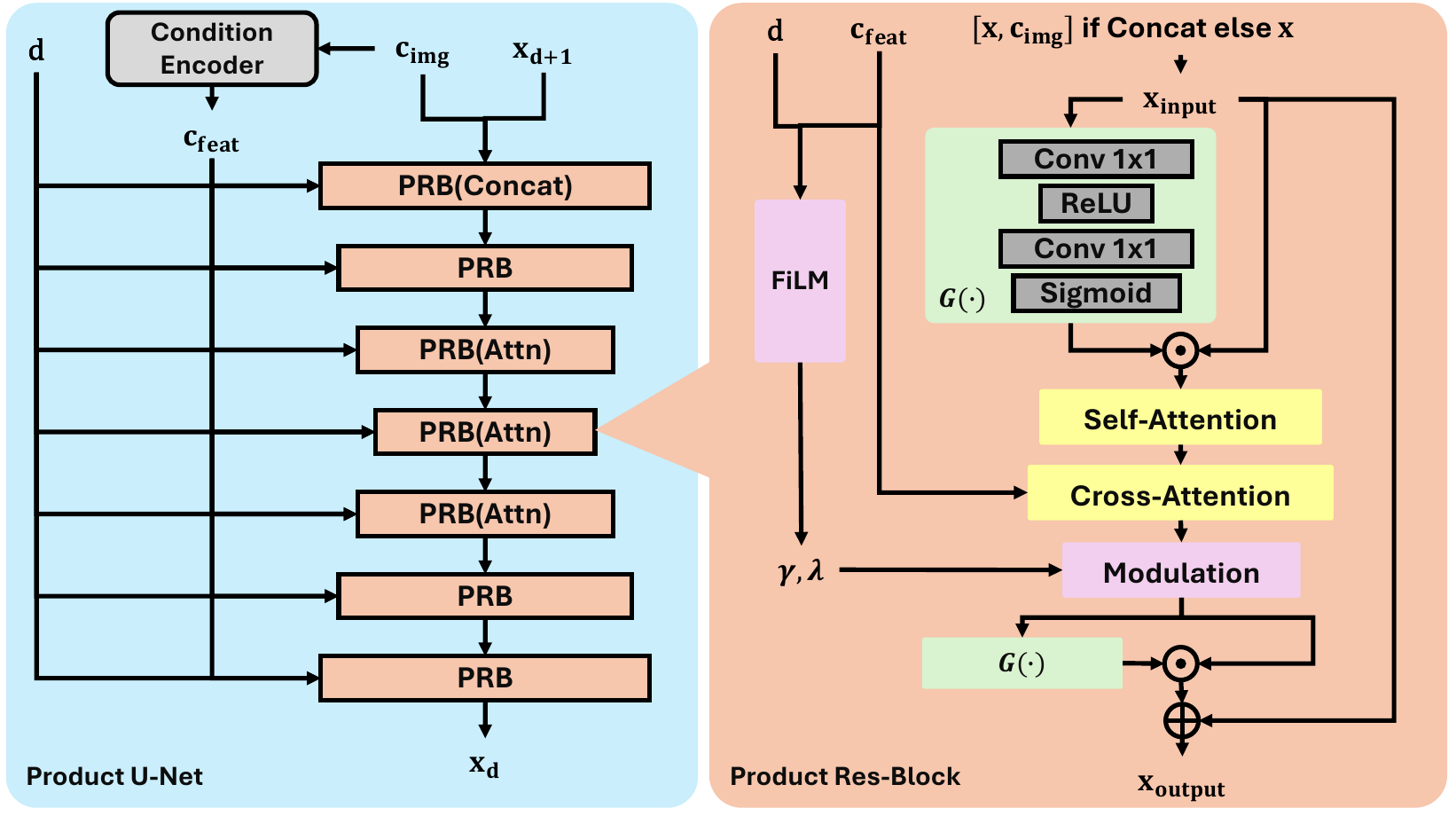}
    %\vspace{-0.1in}
    \caption{Structure of our Product U-Net and Product Res-Block.}
    \label{fig:punet}
    %\vspace{-0.2in}
\end{figure*}

To address this issue, we leverage the convolution theorem, which states that convolution in the spatial domain corresponds to element-wise multiplication in the spectral domain and vice versa. Inspired by this relationship, we design a full-product U-Net that performs input-dependent element-wise modulation in the spatial domain, thereby providing an operator-level approximation to frequency-selective convolutional filtering without explicitly operating in the frequency domain. The architecture is illustrated in Fig.~\ref{fig:punet}. 
Specifically, we introduce a spatial full-product layer:
\begin{equation}
    %\small
    \mathbf{h} = \mathbf{x}_{input}\odot\mathbf{w} = \mathbf{x}_{input}\odot G(\mathbf{x}_{input}),~\label{eq:product_layer}
\end{equation}
where $\odot$ denotes the element-wise multiplication, and $G(\cdot)$ is implemented using a $1\times 1$ convolution with a bottleneck structure. Unlike traditional convolutional layers with static kernels, the modulation weight $\mathbf{w}$ is dynamically generated conditioned on the input features. This allows adaptive feature modulation while significantly reducing computational cost. The bottleneck further improves efficiency by compressing intermediate channels. 

The computational reduction of the proposed full-product U-Net can be quantified by comparing the FLOPs of a standard convolutional layer and the proposed product layer. Since we retain components such as self-attention, cross-attention, and other activation functions, the primary architectural difference lies in replacing standard convolutional layers with full-product layers. Suppose the input feature has size $(C_{in}, H, W)$ and the output feature has size $(C_{out}, H, W)$. For a commonly used $3\times3$ convolutional layer, the numbers of additions and multiplications are both
\begin{equation}
    %\small
    3 * 3 * C_{in} * C_{out} * H * W.~\label{eq:33conv_add_multi}
\end{equation}
For simplicity, let $C_{in} = C_{out} = C$, then the total FLOPs of a single convolutional layer are 
\begin{equation}
    %\small
    2 * (3 * 3 * C * C * H * W) = 18C^{2}HW.~\label{eq:33conv_total}
\end{equation}
In contrast, our product layer employs two consecutive $1\times1$ convolutional layers with a bottleneck ratio $r_{C}$, the total FLOPs of  these two convolutions are
\begin{equation}
    %\small
    \begin{aligned}
        2 * (1 * 1 * (C * \frac{C}{r_{C}} + \frac{C}{r_{C}} * C) * H * W) = \frac{4}{r_{C}}C^{2}HW.~\label{eq:prod_conv}
    \end{aligned}
\end{equation}
Taking into account the element-wise multiplication between $\mathbf{w}$ and $\mathbf{x}_{input}$, whose FLOPs are $CHW$, the total number of operations within the product layer becomes
\begin{equation}
    %\small
    \frac{4}{r_{C}}C^{2}HW + CHW.~\label{eq:prod_total}
\end{equation}
Typically, we choose $r_{C}=4$. In this case, the theoretical per-layer computation reduction for a single layer replacement is about
\begin{equation}
    %\small
    \frac{18C^{2}HW}{C^{2}HW + CHW} = \frac{18C}{C + 1} \overset{\text{large}\,C}{\approx} 18,~\label{eq:speedup}
\end{equation}
While the end-to-end runtime improvement depends on implementation details and hardware, this layer-wise analysis highlights the computational advantage of the product formulation. 

Importantly, although the spectral perturbation is defined in Sec.~\ref{sec:3.2}, the denoiser can now operate entirely in the spatial domain. Instead of operating on complex-valued noise $\mathbf{\epsilon}_{f}$ in the spectral domain, we realize the same structured perturbation in the spatial domain through its induced counterpart. Specifically, once masked noise $\mathbf{M}_{d}\odot\mathbf{\epsilon}_{f}$ is injected in the spectral domain, the corresponding spatial perturbation is uniquely determined by the inverse transform:
\begin{equation}
    %\small
    \mathbf{\epsilon}_{s} \triangleq \mathcal{F}^{-1}(\mathbf{M}_{d}\odot\mathbf{\epsilon}_{f}).~\label{eq:spatial_noise_induce}
\end{equation}
We train the denoiser to predict this induced spatial perturbation directly:
\begin{equation}
    %\small
    \mathbf{\epsilon}_{s} = \mathbf{\epsilon}_{\phi}(\mathbf{x}_{s, d}, d, \mathbf{c}_{\text{img}}),~\label{eq:spatial_noise}
\end{equation}
given diffusion step $d$ and the conditional image $\mathbf{c}_{\text{img}}$ (i.e., the input rainy image $\mathbf{O}$). $\mathbf{\epsilon}_{s}$ is not an additional modeling assumption, instead, it is the deterministic image-domain perturbation produced by the masked spectral noise we sample during training. Thus, we can follow the standard noise-prediction training paradigm while keeping the perturbation design in the spectral domain and the denoiser in the spatial domain. 

Since FFT and IFFT are linear unitary transforms, the spectral forward update in Eq.~\ref{eq:spectral_diff_forward_1} induces an equivalent recursion in the spatial domain. Let $\mathbf{x}_{s, d} = \mathcal{F}^{-1}(\mathbf{x}_{f, d})$ denote the spatial-domain image, then
\begin{equation}
    %\small
    \begin{aligned}
        \mathbf{x}_{s, d} &= \mathcal{F}^{-1}(\mathbf{x}_{f, d}) \\
        &= \mathcal{F}^{-1}\left(\sqrt{1 - \beta_{d}}\mathbf{x}_{f, d - 1} + \sqrt{\beta_{d}}(\mathbf{M}_{d}\odot\mathbf{\epsilon}_{f})\right) \\
        &= \sqrt{1 - \beta_{d}}\mathcal{F}^{-1}(\mathbf{x}_{f, d - 1}) + \sqrt{\beta_{d}}\mathcal{F}^{-1}(\mathbf{M}_{d}\odot\mathbf{\epsilon}_{f}) \\
        &= \sqrt{1 - \beta_{d}}\mathbf{x}_{s, d - 1} + \sqrt{\beta_{d}}\mathbf{\epsilon}_{s}.~\label{eq:fs_transform}
    \end{aligned}
\end{equation}
In other words, the same noise schedule $\beta_{d}$ can be used consistently in both spectral and spatial domains, while the structured perturbation effect is carried by $\mathbf{\epsilon}_{s}$. During training, as shown in Algorithm.~\ref{alg:training}, we sample $\mathbf{\epsilon}_{f}$, apply the mask $\mathbf{M}_{d}$ in the spectral domain, obtain $\mathbf{\epsilon}_{s}$ via Eq.~\ref{eq:spatial_noise_induce}, and optimize $\mathbf{\epsilon}_{\phi}(\cdot)$ to predict $\mathbf{\epsilon}_{s}$ from $\mathbf{x}_{s, d}$. In implementation, the IFFT output is used as the spatial-domain sample $\mathbf{x}_{s,d}$ under standard numerical precision. Note that the inversion formula used in Algorithm.~\ref{alg:training} is algebraically equivalent to Eq.~\ref{eq:spatial_noise_induce}, and is adopted to ensure numerical consistency with the realized noisy sample $\mathbf{x}_{s,d}$.

%\vspace{-0.25in}
\begin{algorithm}[!htpb]
    %\small
    \caption{Training Process of SpectralDiff}\label{alg:training}
    \textbf{Input}: Dataset of clean and rainy image pairs $\mathcal{D} = \{\mathbf{B}_{i}, \mathbf{O}_{i}\}_{i=1}^{|\mathcal{D}|}$ \\
    \textbf{Parameter}: Total diffusion step $D$, noise schedule $\{\beta_{d}\}_{d=1}^{D}$, pre-computed modulation mask $\{\mathbf{M}_{d}\}_{d=1}^{D}$ \\
    \textbf{Output}: Parameters $\phi$ of SpectralDiff
    \hrule
    \begin{algorithmic}[1]
        \REPEAT
        \STATE Randomly select a paired sample $\{\mathbf{B}_{i}, \mathbf{O}_{i}\}$
        \begin{itemize}
            \item[] Let the clean image $\mathbf{x}_{s, 0} = \mathbf{B}_{i}$
            \item[] Let the condition image $\mathbf{c}_{\text{img}} = \mathbf{O}_{i}$
        \end{itemize}
        \STATE Randomly sample a step $d \sim \text{Uniform}\{1, ..., D\}$.
        \begin{itemize}
            \item[] Let $\bar{\alpha}_{d} = \prod_{t=1}^{d}(1 - \beta_{t})$
            \item[] Use the corresponding pre-computed mask $\mathbf{M}_{d}$ for step $d$
        \end{itemize}
        \STATE Add structured noise in the spectral domain
        \begin{itemize}
            \item[] Convert the clean image to the spectral domain \\ {\centering $\mathbf{x}_{f, 0} = \mathcal{F}(\mathbf{x}_{s, 0})$ \par}
            \item[] Sample complex Gaussian noise \\ {\centering $\mathbf{\epsilon}_{f} \sim \mathcal{N}_{\mathbb{C}}(\mathbf{0}, \mathbf{I})$ \par}
            \item[] Inject masked noise \\ {\centering $\mathbf{x}_{f, d} = \sqrt{\bar{\alpha}_{d}}\mathbf{x}_{f, 0} + \sqrt{1 - \bar{\alpha}_{d}}(\mathbf{M}_{d}\odot\mathbf{\epsilon}_{f})$ \par}
        \end{itemize}
        \STATE Compute induced spatial perturbation
        \begin{itemize}
            \item[] Convert back to the spatial domain \\ {\centering $\mathbf{x}_{s, d} = \mathcal{F}^{-1}(\mathbf{x}_{f, d})$ \par}
            \item[] Define the induced spatial noise by inversion \\ {\centering $\mathbf{\epsilon}_{s} = (\mathbf{x}_{s, d} - \sqrt{\bar{\alpha}_{d}}\mathbf{x}_{s, 0}) / \sqrt{1 - \bar{\alpha}_{d}}$ \par}
            %\item[] (Discard negligible imaginary numerical residue to obtain the real-valued noisy sample, ensuring consistency with $\mathbf{x}_{s,d}$.)
        \end{itemize}
        \STATE Estimate the spatial noise with the denoiser \\ {\centering $\hat{\mathbf{\epsilon}}_{s} = \mathbf{\epsilon}_{\phi}(\mathbf{x}_{s, d}, d, \mathbf{c}_{\text{img}})$ \par}
        \STATE Minimize the noise-prediction objective (MSE) \\ {\centering $\mathcal{L}(\phi) = \left\lVert\mathbf{\epsilon}_{s} - \hat{\mathbf{\epsilon}}_{s}\right\rVert_{2}^{2}$ \par}
        \STATE Update parameters $\phi$ by gradient descent on $\mathcal{L}(\phi)$
        \UNTIL{converged}
        \STATE \textbf{return} $\phi$
    \end{algorithmic}
\end{algorithm}
%\vspace{-0.25in}

During inference, we start from $\mathbf{x}_{s, D} = \mathbf{O}$ and adopt a deterministic DDIM-style update rule as a parameterized conditional deraining trajectory under the structured corruption, rather than interpreting it as the exact posterior associated with the forward kernel. Instead of using uniformly spaced DDIM steps, we estimate a directional probability distribution $p(\theta)$ from the frequency spectrum of the input rainy image and sample denoising steps accordingly. This non-uniform scheduling allocates more inference budget to dominant rain directions and scales, improving both inference efficiency and directional consistency in the reconstructed image. Detailed inference procedure is provided in Algorithm.~\ref{alg:testing}.

%\vspace{-0.25in}
\begin{algorithm}[!htpb]
    %\small
    \caption{Inference Process of SpectralDiff}\label{alg:testing}
    \textbf{Input}: A rainy image $\mathbf{O}$ \\
    \textbf{Parameter}: Total diffusion steps $D$, number of DDIM steps $S$, noise schedule $\{\beta_{d}\}_{d=1}^{D}$, SpectralDiff denoiser $\mathbf{\epsilon}_{\phi}(\cdot)$ \\
    \textbf{Output}: A clean image $\mathbf{B}$
    \hrule
    \begin{algorithmic}[1]
        \STATE Estimate directional probability $p(\theta)$ from $\mathbf{O}$
        \STATE Assign each step $d \in \{1, ..., D\}$ a weight $w_d \propto  p(\theta_{d})$ 
        \STATE Sample $(S-1)$ steps with replacement according to $\{w_d\}_{d=1}^{D}$ and include $D$: \\
        {\centering $\{d_1, \dots, d_{S-1}\} \sim \text{Multinomial}(w_1, \dots, w_D)$, \quad $d_S \leftarrow D$ \par}
        \STATE Sort the steps in non-increasing order: $d_S \ge d_{S-1} \ge \dots \ge d_1$
        \STATE Pre-compute $\bar{\alpha}_d = \prod_{t=1}^{d}(1 - \beta_t)$
        \STATE Initialize $\hat{\mathbf{x}}_{s, d_S} = \mathbf{O}$
        \STATE Let the condition image $\mathbf{c}_{\text{img}} = \mathbf{O}$
        \FOR{$i = S, ..., 1$}
            \STATE Estimate current noise \\ {\centering $\hat{\mathbf{\epsilon}}_{s} = \mathbf{\epsilon}_{\phi}(\hat{\mathbf{x}}_{s, d_i}, d_i, \mathbf{c}_{\text{img}})$ \par}
            \STATE Estimate clean image at step $d_i$ \\ {\centering $\hat{\mathbf{x}}_{s, 0} = (\hat{\mathbf{x}}_{s, d_i} - \sqrt{1 - \bar{\alpha}_{d_i}}\,\hat{\mathbf{\epsilon}}_{s}) / \sqrt{\bar{\alpha}_{d_i}}$ \par}
            \IF{$i > 1$}
            \STATE Deterministic DDIM update \\
            {\centering $\hat{\mathbf{x}}_{s, d_{i-1}} = \sqrt{\bar{\alpha}_{d_{i-1}}}\hat{\mathbf{x}}_{s, 0} + \sqrt{1 - \bar{\alpha}_{d_{i-1}}}\hat{\mathbf{\epsilon}}_{s}$ \par}
            \ENDIF
        \ENDFOR
        \STATE \textbf{return} $\mathbf{B} = \hat{\mathbf{x}}_{s, 0}$
    \end{algorithmic}
\end{algorithm}
%\vspace{-0.4in}

%--------------------------------------------------
%       Section 4 : Experiments
%--------------------------------------------------

% ~ 2 page

%\newpage
\section{Experiments~\label{sec:4}}

%%%% Section 4.1 : Datasets
%%%%%%%%%%%%%%%%%%%%%%%%%%%
\subsection{Datasets~\label{sec:4.1}}

We evaluate on \textbf{Rain1400~\cite{fu2017removing}}, \textbf{RainCityscapes~\cite{hu2019depth}}, and \textbf{SPA-Data~\cite{wang2019spatial}}, covering both synthetic paired settings and real-world rain scenarios. Detailed statistics and data splits are provided in Appendix.~\ref{sec:4.1_1}.

%%%% Section 4.2 : Evaluation Metrics
%%%%%%%%%%%%%%%%%%%%%%%%%%%%%%%%%%%%%
%\input{Tex/sec4_2}

%%%% Section 4.3 : Implementation Details
%%%%%%%%%%%%%%%%%%%%%%%%%%%%%%%%%%%%%%%%%
%\input{Tex/sec4_3}

%%%% Section 4.4 : Overall Performance
%%%%%%%%%%%%%%%%%%%%%%%%%%%%%%%%%%%%%%

\begin{table*}[!htpb]
    %%\vspace{-0.25in}
    \centering
    \renewcommand\arraystretch{0.8}
    %\small
    \caption{Overall performance (PSNR ($\uparrow$), SSIM ($\uparrow$)) and efficiency (Time[$s^{\dagger}$] ($\downarrow$)) comparison among different deraining methods.}
    \label{tab:overall}
    %\vspace{-0.1in}
    \begin{tabular}{c c c c c c c c c c}
        \toprule
            &\multicolumn{3}{c}{Rain1400} &\multicolumn{3}{c}{RainCityscapes} &\multicolumn{3}{c}{SPA-Data} \\
        \cmidrule(lr){2-4}\cmidrule(lr){5-7}\cmidrule(lr){8-10}
            &PSNR &SSIM &Time &PSNR &SSIM &Time &PSNR &SSIM  &Time \\
        \midrule
            DSC~\cite{luo2015removing} &27.60 &0.825 &0.320 &19.90 &0.795 &0.341 &24.26 &0.696 &0.321 \\ 
            LayerPrior~\cite{li2016rain} &24.75 &0.780 &7.398 &13.71 &0.736 &4.461 &33.03 &0.867 &5.065 \\ 
            DDN~\cite{fu2017removing} &28.33 &0.815 &0.001 &21.99 &0.819 &0.001 &34.06 &0.869 &0.002 \\ 
            PReNet~\cite{ren2019progressive} &33.40 &0.901 &0.007 &28.56 &0.902 &0.007 &35.55 &0.886 &0.014 \\ 
            SIRR~\cite{wei2019semi} &30.70 &0.874 &0.002 &22.08 &0.846 &0.002 &36.39 &0.891 &0.002 \\ 
            DiG-CoM~\cite{ran2020single} &25.19 &0.772 &0.466 &13.78 &0.734 &0.297 &33.62 &0.863 &0.378 \\
            RCDNet~\cite{wang2020model} &26.59 &0.795 &0.071 &22.99 &0.887 &0.083 &31.55 &0.867 &0.054 \\
            DRCDNet~\cite{wang2023rcdnet} &30.97 &0.849 &0.356 &25.09 &0.892 &0.375 &36.40 &0.894 &0.280 \\
            WeatherDiff~\cite{ozdenizci2023restoring} &30.28 &0.855 &3.117 &23.58 &0.889 &3.419 &36.86 &0.894 &6.606 \\
            RainDiff~\cite{wei2023raindiffusion} &25.93 &0.791 &11.159 &20.49 &0.850 &10.604 &30.69 &0.851 &9.676 \\
            DM-DIRR~\cite{zeng2024multi} &20.21 &0.491 &4.169 &19.66 &0.689 &4.630 &23.53 &0.684 &2.306 \\
        \midrule
            \textbf{SpectralDiff} &30.39 &0.860 &0.114 &29.04 &0.915 &0.115 &38.03 &0.895 &0.118 \\
        \bottomrule
    \end{tabular}
    %\vspace{-0.2in}
\end{table*}
\footnotetext{$^{\dagger}$ Time is measured under each method's \emph{author-recommended} inference settings (e.g., WeatherDiff, RainDiff, DM-DIRR: $100$ steps, SpectralDiff: 10 steps)}

\subsection{Overall Performance~\label{sec:4.4}}

We compare SpectralDiff with representative optimization-based~\cite{luo2015removing, li2016rain, ran2020single}, NN-based~\cite{fu2017removing, ren2019progressive, wang2020model, wang2023rcdnet, wei2019semi}, and diffusion-based methods~\cite{ozdenizci2023restoring, wei2023raindiffusion, zeng2024multi}. 
To ensure a controlled comparison, all baselines are retrained on each dataset following their originally reported training protocols, with the batch size standardized across methods (details in Appendix.~\ref{sec:4.3}). PSNR and SSIM are used as evaluation metrics. As shown in Table.~\ref{tab:overall}, SpectralDiff achieves competitive performance on synthetic benchmarks and superior performance on real-world SPA-Data. On synthetic datasets such as Rain1400 and RainCityscapes, several spatial-domain models obtain slightly higher PSNR/SSIM. However, these methods exhibit noticeable performance degradation on real-world SPA-Data, where rain streaks present more complex and diverse structures. In contrast, SpectralDiff maintains stable performance across both synthetic and real settings, suggesting improved robustness to natural rain distributions. Table.~\ref{tab:overall} also reports average inference time measured under each method’s author-recommended inference configuration. These diffusion baselines follow a conditional generation paradigm starting from pure noise, while SpectralDiff adopts a denoising-based trajectory initialized from the rainy image, which is more aligned with image restoration settings. Diffusion-based baselines follow their recommended $100$-step settings. In contrast, SpectralDiff converges within $10$ steps, resulting in substantially lower latency while maintaining competitive fidelity. Even when normalized by the number of steps, SpectralDiff exhibits lower average per-step runtime than diffusion baselines, reflecting the efficiency of the product U-Net design. 
%Detailed implementation settings are provided in Appendix.~\ref{sec:code} to facilitate reproducibility, and the complete code will be publicly released upon acceptance.

%%%% Section 4.5 : Ablation Study
%%%%%%%%%%%%%%%%%%%%%%%%%%%%%%%%%
\subsection{Ablation Study~\label{sec:4.5}}

We conduct ablation studies to evaluate the impact of the spectral perturbation design and the product U-Net architecture.

\paragraph{\textbf{Spatial \textit{vs.} Spectral Diffusion.}} 
We compare spatial diffusion and spectral diffusion under identical architectures and training settings. Both variants adopt the same denoising-based formulation initialized from the rainy image, ensuring that the comparison isolates the effect of domain and perturbation design. As shown in Table.~\ref{tab:spatial_spectral}, naive spectral variants (magnitude-only, phase-only, or unmasked spectral noise) suffer from severe degradation, indicating that frequency representation alone is insufficient. In contrast, introducing the structured mask $\mathbf{M}_d$ restores performance to a level comparable to spatial diffusion while adopting a frequency-aware perturbation mechanism. This demonstrates that the performance gain arises from the structured spectral corruption rather than from simply switching domains.
\begin{table}[!htpb]
    %\vspace{-0.2in}
    \centering
    \renewcommand\arraystretch{0.9}
    %\small
    \caption{Performance comparison of diffusion models deraining in spatial domain and spectral domain on Rain1400.}
    \label{tab:spatial_spectral}
    %\vspace{-0.1in}
    \begin{tabular}{c c c c c}
        \toprule
            Diffusion ($S=10$) &PSNR &SSIM &FLOPs &Params \\
        \midrule
            Spatial ($\mathbf{\epsilon}_{s}$) &30.93 &0.853 &376.60 &6.08 \\ 
            Spectral ($\lvert\mathbf{x}_{f}\rvert$) &6.89 &0.114 &376.60 &6.08 \\ 
            Spectral ($\angle\mathbf{x}_{f}$) &24.27 &0.806 &376.60 &6.08  \\ 
            Spectral ($\mathbf{\epsilon}_{f}$) &14.68 &0.246 &1269.59 &16.66 \\ 
            Spectral ($\mathbf{M}_{d}\odot\mathbf{\epsilon}_{f}$) &30.86 &0.857 &1269.59 &16.66 \\
        \bottomrule
    \end{tabular}
    %\vspace{-0.2in}
\end{table}

\paragraph{\textbf{Convolutional \textit{vs.} Product U-Net.}}
Table.~\ref{tab:convolution_product} compares a standard convolutional U-Net and the proposed product U-Net within the SpectralDiff framework. The product design significantly reduces FLOPs and parameters while incurring only marginal performance change, consistent with the efficiency gains observed during inference. 
\begin{table}[!htpb]
    %\vspace{-0.2in}
    \centering
    \renewcommand\arraystretch{0.9}
    %\small
    \caption{Performance comparison of our SpectralDiff model with different U-Nets on Rain1400.}
    \label{tab:convolution_product}
    %\vspace{-0.1in}
    \begin{tabular}{c c c c c}
        \toprule
           ($S=10$) &PSNR &SSIM &FLOPs &Params \\
        \midrule
            SpectralDiff (Conv) &30.86 &0.857 &1269.59 &16.66 \\
            SpectralDiff (Prod) &30.39 &0.860 &199.81 &3.15 \\
        \bottomrule
    \end{tabular}
    %\vspace{-0.2in}
\end{table}

%--------------------------------------------------
%       Section 5 : Conclusion
%--------------------------------------------------

% ~ 1/4 page

%\newpage
\section{Conclusion~\label{sec:5}}

In this paper, we propose SpectralDiff, a diffusion-based framework for single-image rain removal. By analyzing the layered characteristics of rain streaks in the frequency domain, we introduce a structured spectral perturbation mechanism that enables progressive and frequency-aware rain suppression within a denoising-based diffusion trajectory. To improve computational efficiency, we further replace conventional convolutional layers with a full-product U-Net, leveraging the connection between spectral convolution and spatial element-wise modulation. Experiments on both synthetic and real-world datasets demonstrate that SpectralDiff achieves competitive or superior deraining performance while substantially reducing model complexity and inference latency. Future work will explore adaptive spectral masking and more general structured corruption designs for broader restoration scenarios.

%Bibliography
\bibliographystyle{splncs04}
\bibliography{eccv26} 

\newpage
\appendix
\renewcommand\thesubsection{\thesection\arabic{subsection}}
\section*{Appendix~\label{sec:appendix}}

\subsection{Experimental Settings -- Dataset Details~\label{sec:4.1_1}}

\begin{itemize}
    \item \textbf{Rain1400~\cite{fu2017removing}:} 
    A synthetic benchmark containing $12600$ training images and $1400$ testing images. 
    We further split the training set into training and validation subsets with an $85\% : 15\%$ ratio. 
    Rain streaks are synthesized using a photo-realistic rendering model~\cite{garg2006photorealistic}. 
    Following common practice, all images are resized to $256\times256$ and normalized to $[0,1]$ for both training and evaluation.
    \item \textbf{RainCityscapes~\cite{hu2019depth}:} 
    This dataset consists of $10620$ outdoor images ($9432$ for training and $1188$ for testing) synthesized with more complex rain–background blending compared to Rain1400. 
    During training, we randomly crop $256\times256$ patches, while center cropping is applied at test time. 
    All inputs are normalized to $[0,1]$.
    \item \textbf{SPA-Data~\cite{wang2019spatial}:} 
    A real-world rain dataset constructed from $170$ rain videos ($84$ collected from public sources and $86$ manually captured). 
    The training set contains $638{,}492$ image pairs and the testing set contains $1000$ pairs. 
    Clean targets are estimated based on temporal consistency across video frames as described in~\cite{wang2019spatial}. 
    We apply the same $256\times256$ cropping and normalization strategy as in RainCityscapes.
\end{itemize}

\subsection{Experimental Settings -- Evaluation Metrics~\label{sec:4.2}}

In order to evaluate the deraining effectiveness, we adopt the following metrics, computed on RGB images normalized to $[0, 1]$, to measure the quality of images after rain removal:

\begin{itemize}
    \item \textbf{Peak-Signal-Noise-Ratio (PSNR, $\uparrow$)~\cite{huynh2008scope}:} Given a clean image $\mathbf{B}$ with size of $m\times n$ and a derained image $\hat{\mathbf{B}}$ of the same size, we can calculate the Mean-Squared-Error (MSE) as
    \begin{equation}
        MSE(\mathbf{B}, \hat{\mathbf{B}}) = \frac{1}{mn}\sum_{i=0}^{m-1}\sum_{j=0}^{n-1}[\hat{\mathbf{B}}(i,j)-\mathbf{B}(i,j)]^2~\label{eq:mse}.
    \end{equation}
    Then, PSNR can be defined as
    \begin{equation}
        PSNR(\mathbf{B}, \hat{\mathbf{B}}) = 10\log_{10}(\frac{MAX(\mathbf{B})^2}{MSE(\mathbf{B}, \hat{\mathbf{B}})})~\label{eq:psnr},
    \end{equation}
    where $MAX(\mathbf{B})$ is the maximum pixel value in image $\mathbf{B}$. The larger the PSNR value, the smaller the distortion introduced during rain removal.
    \item \textbf{Structure Similarity (SSIM, $\uparrow$)~\cite{wang2004image}:} SSIM compares clean image $\mathbf{B}$ and derained image $\hat{\mathbf{B}}$ in three aspects: luminance ($l$), contrast ($c$) and structure ($s$) and can be defined as
    \begin{equation}
        SSIM(\mathbf{B}, \hat{\mathbf{B}}) = [l(\mathbf{B}, \hat{\mathbf{B}})^{\alpha}\cdot c(\mathbf{B}, \hat{\mathbf{B}})^{\beta}\cdot s(\mathbf{B}, \hat{\mathbf{B}})^{\gamma}]~\label{eq:ssim},
    \end{equation}
    and $l$, $c$ and $s$ are calculated as
    \begin{equation}
        \begin{aligned}
            l(\mathbf{B}, \hat{\mathbf{B}}) &= \frac{2\mu_{\mathbf{B}}\mu_{\hat{\mathbf{B}}} + c_{1}}{\mu_{\mathbf{B}}^2 + \mu_{\hat{\mathbf{B}}}^2 + c_{1}}, \\
            c(\mathbf{B}, \hat{\mathbf{B}}) &= \frac{2\sigma_{\mathbf{B}}\sigma_{\hat{\mathbf{B}}} + c_{2}}{\sigma_{\mathbf{B}}^2 + \sigma_{\hat{\mathbf{B}}}^2 + c_{2}}, \\
            s(\mathbf{B}, \hat{\mathbf{B}}) &= \frac{\sigma_{\mathbf{B}\hat{\mathbf{B}}} + c_{3}}{\sigma_{\mathbf{B}}\sigma_{\hat{\mathbf{B}}} + c_{3}},~\label{eq:lcs}
        \end{aligned}
    \end{equation}
    where $\mu_{\mathbf{B}}$ and $\mu_{\hat{\mathbf{B}}}$ are mean values of images $\mathbf{B}$ and $\hat{\mathbf{B}}$, $\sigma_{\mathbf{B}}^2$ and $\sigma_{\hat{\mathbf{B}}}^2$ are corresponding variances, $\sigma_{\mathbf{B}\hat{\mathbf{B}}}$ is the covariance between $\mathbf{B}$ and $\hat{\mathbf{B}}$, $c_1=(K_1L)^2$, $c_2=(K_2L)^2$ and $c_3=c_2/2$ are constant values. For simplicity, let $\alpha=\beta=\gamma=1$ and Eq.~\ref{eq:ssim} can be rewritten as
    \begin{equation}
        SSIM(\mathbf{B}, \hat{\mathbf{B}}) = \frac{(2\mu_{\mathbf{B}}\mu_{\hat{\mathbf{B}}} + c_{1})(2\sigma_{\mathbf{B}\hat{\mathbf{B}}} + c_{2})}{(\mu_{\mathbf{B}}^2 + \mu_{\hat{\mathbf{B}}}^2 + c_{1})(\sigma_{\mathbf{B}}^2 + \sigma_{\hat{\mathbf{B}}}^2 + c_{2})}.~\label{eq:ssim_simple}
    \end{equation}
    The larger the SSIM value, the better the image quality after rain removal.
\end{itemize}

\subsection{Experimental Settings -- Implementation Details~\label{sec:4.3}}

All experiments are conducted on a single 48G \texttt{NVIDIA RTX 6000 Ada Generation} GPU. 
Models are optimized using \texttt{AdamW} with a \texttt{ReduceLROnPlateau} scheduler, where the learning rate decreases from $2e-4$ to $1e-6$. A batch size of $4$ is used for training and 1 for evaluation across all methods.

SpectralDiff is trained using DDPM with $1080$ diffusion steps under a cosine noise schedule. 
The number of diffusion steps equals the total number of predefined spectral masks (see below). During inference, we adopt deterministic DDIM sampling with $10$ steps.

The U-Net backbone follows a symmetric encoder–decoder architecture with three down-sampling blocks, one middle block, and three up-sampling blocks. The base channel dimension is $64$, with channel sizes $\{64, 128, 256\}$ in the encoder and $\{128, 64, 64\}$ in the decoder. All blocks are implemented as residual blocks. Self-attention and cross-attention are applied in the last downsampling block, the middle block, and the first upsampling block. The conditional rainy image is concatenated with the input at the first downsampling layer.

Directional spectral masks are constructed by enumerating representative parameter combinations. Specifically, $r_{d}\in\{0.1,0.3,0.5\}$ and $\sigma_{r_{d}}\in\{0.05,0.2\}$ (relative to the maximum frequency radius), $\theta_{d}$ is sampled every $3^{\circ}$ within $[0^{\circ},180^{\circ})$ ($60$ orientations), and $\kappa_{d}\in\{2,5,10\}$. All combinations of $(r_{d}, \sigma_{r_{d}}, \theta_{d}, \kappa_{d})$ result in $1080$ predefined masks, each assigned to a corresponding diffusion step. These masks are fixed across all datasets and serve as a global spectral prior rather than dataset-specific hyperparameters.

\end{document}